\documentclass[letterpaper]{article} 
\usepackage[preprint]{aaai2027}
\usepackage[hyphens]{url}  
\usepackage{graphicx} 
\usepackage{natbib}  
\usepackage{caption} 
\usepackage{algorithm}
\usepackage{algorithmic}
\usepackage{bm}
\usepackage{amsthm,amsmath}
\usepackage{amssymb}
\usepackage[table]{xcolor} 
\definecolor{lightgray}{gray}{0.92} 
\usepackage{diagbox}
\usepackage{subcaption}
\usepackage{multirow}
\usepackage{makecell}

\usepackage{newfloat}
\usepackage{listings}
\DeclareCaptionStyle{ruled}{labelfont=normalfont,labelsep=colon,strut=off} 
\floatstyle{ruled}
\newfloat{listing}{tb}{lst}{}
\floatname{listing}{Listing}

\usepackage{booktabs}

\title{WeCon: An Efficient Weight-Conditioned Neural Solver for Multi-Objective Combinatorial Optimization Problems}
\author{
    Xuan Wu\textsuperscript{\rm 1}, Jinbiao Chen\textsuperscript{\rm 2}, Yang Li\textsuperscript{\rm 3}, Lijie Wen\textsuperscript{\rm 4}, Chunguo Wu\textsuperscript{\rm 1}, Yuanshu Li\textsuperscript{\rm 1,5}, Yubin Xiao\textsuperscript{\rm 1}, Chunyan Miao\textsuperscript{\rm 6}, You Zhou\textsuperscript{\rm 1}\corresponding, Di Wang\textsuperscript{\rm 6}
}
\affiliations{
    \textsuperscript{\rm 1} 
College of Computer Science and Technology, Jilin University, China\\
    \textsuperscript{\rm 2} Department of Industrial Systems Engineering and Management, National University of Singapore, Singapore \\
     \textsuperscript{\rm 3} College of Software, Jilin University, China \\
    \textsuperscript{\rm 4} School of Software, Tsinghua University, China \\
    \textsuperscript{\rm 5} School of Computing and Information Systems, Singapore Management University, Singapore \\
    \textsuperscript{\rm 6} LILY Research Centre,
Nanyang Technological University, Singapore
    
}

\begin{document}

\maketitle

\begin{abstract}
Existing neural solvers for Multi-Objective Combinatorial Optimization Problems (MOCOPs) commonly adopt decomposition-based strategies that scalarize a MOCOP into multiple subproblems associated with distinct weight vectors. However, they either inject weights only once during decoding, limiting weight-conditioned context modeling, or primarily during encoding, causing weight-signal dilution during decoding. Moreover, their preference optimization methods rely on purely random sampling to construct solution pairs for training solvers, which often produces less informative pairs and thus leads to low training effectiveness. To better address these limitations, we propose an efficient Weight-Conditioned neural solver (\textbf{WeCon}). Specifically, we design an encoder layer with three attention blocks and our proposed Gated Residual Fusion block to facilitate harmonious interaction between instance features and weights, thereby generating informative weight-conditioned context. We further introduce a plug-and-play Residual Fusion block in the decoder to alleviate weight-signal dilution. Finally, we propose Efficient Preference Optimization, which constructs high-quality solutions, thereby generating more informative pairs to improve training effectiveness.  Experimental results on four MOCOP variants across different problem scales and distribution patterns demonstrate that WeCon achieves HyperVolume (HV) performance comparable to the state-of-the-art (SOTA) solver POCCO-W, while requiring approximately 40\% less inference time. Moreover, the variant WeCon-CCO, which adopts an enhanced decoder, achieves the best overall HV performance with increased inference time. Ablation studies validate the contributions of all proposed designs. 
\end{abstract}


\section{Introduction}
\label{sec1}

Combinatorial Optimization Problems (COPs) are at the core of mathematical optimization, aiming to find optimal solutions in the discrete search space, and have long attracted sustained research attention due to their broad real-world relevance \cite{schrijver_history_2005,hercules}. Conventional algorithms for solving COPs can be generally divided into three categories, namely exact, approximation, and heuristic methods \cite{kwon_pomo_2020,survey}; however, most cannot derive insights from historical COP instances, leading to substantial computing overhead \cite{xiao_distilling_2024,10737904}.

To solve COPs efficiently, an emerging number of recent studies developed neural solvers that learn from extensive historical instances, enabling effective searches for optimal solutions \cite{mingzhao,dgl,liyuanshu,xiao2025}. Despite this progress, existing studies have largely focused on Single-Objective COPs (SOCOPs), leaving Multi-Objective COPs (MOCOPs) relatively underexplored. However, many real-world decisions must balance multiple considerations (e.g., cost and convenience). Therefore, developing neural solvers tailored to MOCOPs is of paramount importance \cite{chen2023efficient}.

\begin{figure}[!t]
    \centering
    \includegraphics[width=1\linewidth]{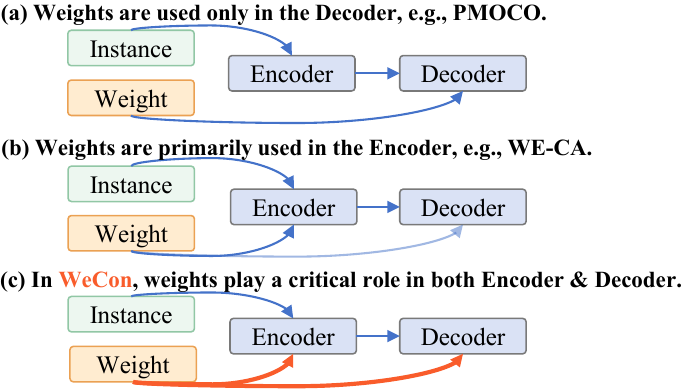}
    \vspace{-0.6cm}
    \caption{Comparative illustration of our proposed WeCon and representative NCO models for solving MOCOPs.}
    \label{fig:com}
    \vspace{-0.6cm}
\end{figure}

To tackle the challenge posed by MOCOPs, recent neural solvers \cite{chen2025image, CNH}  first decompose a MOCOP into a set of subproblems associated with distinct weight vectors, and then train a single solver that generalizes across a wide range of weights. Such a solver, termed a single-model solver, can produce solutions that dynamically adapt to the input weights, i.e., it \textit{enables weight-conditioned decision-making}.
 As shown in Figure~\ref{fig:com}(a) and \ref{fig:com}(b), existing single-model solvers typically incorporate weights either only in the decoder (e.g., PMOCO \cite{lin_pareto_2022}) or primarily in the encoder (e.g., WE-CA \cite{WE}). These two designs, while straightforward, can hinder generalization across weights. Specifically, decoding-only injection may fail to provide informative weight-conditioned context, whereas encoding-primarily injection may dilute the weight signals during decoding. To boost performance, Fan et al. (\citeyear{POCCO}) proposed the Mixture-of-Experts (MoE)-based Conditional Computation (CCO) block in the decoder to route subproblems to different experts. Nevertheless, this gating-and-routing process introduces additional runtime (see Section~\ref{sec5.1}), which may limit model practicality in time-sensitive scenarios, e.g., traffic signal control \cite{10974678}. This inefficiency reflects that such design does not fundamentally address the key limitation of existing solvers, i.e., the underexploitation of weights in the encoder and decoder. Therefore, in this work, we investigate the following research question:

\noindent \textit{Can a solver achieve high-level performance by effectively producing informative weight-conditioned context and preventing weight-signal dilution during decoding, without significantly increasing runtime?}

Moreover, the recent study \cite{PO} proposed  Preference Optimization (PO) to train the solver. Specifically, PO randomly samples a number of $r$ candidate solutions for each instance under the current policy and performs pairwise comparisons based on their objective values to construct $\frac{r(r-1)}{2}$ number of preference pairs, where each pair consists of a better solution $\pi_w$ and a worse solution $\pi_l$. These pairs then provide comparative supervision for training the solver. However, purely random sampling does not guarantee a sufficient number of \mbox{(near-)}optimal solutions are sampled to serve as $\pi_w$, which limits PO’s exploitation ability and adversely affects training effectiveness \cite{BOPO}.

To better overcome the afore-discussed limitations holistically, we propose a Weight-Conditioned neural solver (\textbf{WeCon}). As shown in Figure~\ref{fig:com}(c), WeCon lets weight vectors play an essential, critical role in both encoder and decoder, aiming to achieve high-level performance by sharing weight-conditioned context between them. Specifically, the encoder comprises stacked layers that alternately apply a Multi-Head Self-Attention (MHSA) block over instance features and then use two Multi-Head Attention (MHA) blocks together with our proposed Gated Residual Fusion (GRF) block to integrate instance and weight embeddings (see Section~\ref{sec4.1}), thereby producing informative weight-conditioned context. In the decoder, WeCon adopts the proposed plug-and-play Residual Fusion (RF) block as its core (see Section~\ref{sec4.2}). Specifically, the decoder builds upon the MHA layer that constitutes the entire decoder of WE-CA~\cite{WE}, with the proposed RF module appended immediately after. By adopting this design, WeCon mitigates weight-signal dilution and achieves high-level performance without significantly increasing runtime (see Section~\ref{sec5.1}). 
\textit{The incorporation of RF in the decoder distinguishes WeCon from existing solvers} that primarily utilize weights in the encoder (see Figure~\ref{fig:com}(b)).
For instance, WE-CA treats the weight embedding as a virtual node appended to the instance embeddings along the node dimension, and then uses the concatenated embeddings as the key/value inputs to the decoder’s MHA layer. In contrast, WeCon's decoder first leverages the weight embedding within the MHA layer and subsequently broadcasts it to all nodes, upon which RF explicitly conditions each node embedding according to the weights. To the best of our knowledge, \textbf{WeCon is the first neural solver to simultaneously achieve SOTA performance and runtime efficiency, by effectively leveraging weights in both encoder and decoder.}
Moreover, to further demonstrate the generality of RF, we propose a variant, \textbf{WeCon-CCO}, which adopts the decoder from POCCO-W \cite{POCCO} and augments it with the RF module.
To improve training effectiveness, we extend PO and propose  Efficient Preference Optimization (EPO). Specifically, instead of randomly sampling $r$ solutions for each instance, EPO performs a guided sampling to generate $\lceil \frac{r}{c} \rceil $ solutions, while the remaining $(r-\lceil \frac{r}{c} \rceil)$ solutions are sampled at random. This guided sampling constrains each decision step to select the next node only from the top-$k$ feasible nodes with the highest probabilities (see Section~\ref{sec4.3}). This design enables EPO to obtain sufficiently high-quality solutions, thereby constructing more informative preference pairs with larger quality gaps so as to improve training effectiveness. 


The key contributions of this work are as follows.
 
\textbf{I)} We design an encoder in which each layer applies self-attention over instance features, then performs bidirectional attention between instance and weight features, and finally integrates them via the proposed GRF module to produce weight-conditioned context.

\textbf{II)} We develop a plug-and-play RF block for different decoder architectures to mitigate weight-signal dilution and enable more effective weight-conditioned decision-making.

\textbf{III)} To improve training effectiveness, we propose EPO, which efficiently generates sufficiently high-quality solutions and thereby constructs more informative preference pairs.

\textbf{IV)} To evaluate the effectiveness of the proposed WeCon and WeCon-CCO, we conduct extensive experiments on four MOCOPs across different problem scales and distribution patterns. The experimental results demonstrate that WeCon achieves HV performance comparable to the SOTA solver POCCO-W, while requiring approximately 40\% less inference time. \mbox{WeCon-CCO} attains the best overall HV performance, with the cost of increased inference time. Ablation studies validate the effectiveness of all proposed designs.

\begin{figure*}
    \centering
\includegraphics[width=1\linewidth]{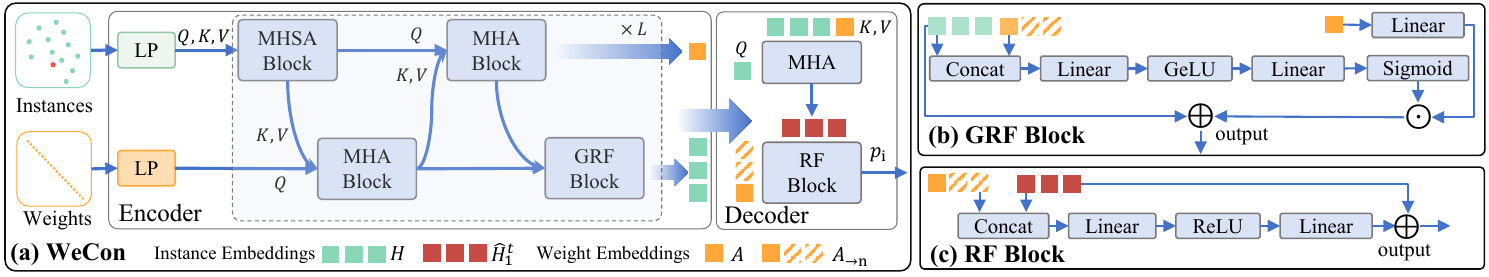}
\vspace{-0.6cm}
    \caption{WeCon network architecture and components. (a) Illustration of the proposed WeCon tailored for MOCOPs. Through multiple rounds of interactions between instance features and weight vectors, WeCon's encoder produces informative weight-conditioned context. Subsequently,  WeCon's decoder exploits the proposed RF block to mitigate weight-signal dilution and enable effective weight-conditioned decisions. (b) and (c) Illustrations of the proposed GRF and RF modules, respectively.}
    \label{fig:WeCon}
    \vspace{-0.4cm}
\end{figure*}
\section{Related Work}
\label{sec2}
In this section, we review the relevant literature.

\noindent\textbf{Neural Solvers for MOCOP:}
To solve MOCOPs, pioneering studies \cite{Wu_2020_MODRLDAM, Zhang_meta_2022} adopted the MOEA/D framework \cite{zhang_MOEA_2007}, which decomposes a MOCOP into a set of subproblems associated with different weight vectors and trains a separate model for each subproblem. Such methods are termed multi-model solvers. Nonetheless, training multiple models requires substantial computational resources. To overcome this limitation, \citeauthor{lin_pareto_2022} (\citeyear{lin_pareto_2022}) proposed a single-model solver that incorporates the weight vector into the decoder, generating solutions that dynamically adapt to different weight vectors. Subsequently, single-model solvers have become the mainstream for solving MOCOPs \cite{pan,chen2025image}.  However, existing solvers often leverage weight vectors coarsely, either primarily in the encoder or solely in the decoder (see Figure~\ref{fig:com}), which may dilute the weight signals during decoding or fail to provide informative weight-conditioned context. 


\noindent \textbf{Preference Optimization:} Most neural solvers are trained with Reinforcement Learning (RL), while fewer adopt Supervised Learning (SL) \cite{kwon_pomo_2020,survey}. Because RL does not require optimal solutions as training labels, adopting RL substantially reduces labeling costs \cite{wu2026efficient}. However, RL methods, such as REINFORCE, update the policy by comparing sampled rewards against a baseline. As training progresses, the policy-gradient signal may diminish, leading to slow convergence. To mitigate this problem,  \citeauthor{PO} (\citeyear{PO}) proposed PO as an alternative to RL, which samples multiple solutions for each instance and then constructs preference pairs based on their relative objective values, enabling the solver to learn comparative quality across solutions. Subsequently, to improve the training effectiveness of PO, BOPO \cite{BOPO} and VAGPO \cite{vagpo}  select only a subset of the sampled solutions to construct preference pairs, instead of exhaustively considering all sampled solutions. However, these methods largely overlook the sampling procedure itself, resulting in limited improvement. In this work, we propose a more effective PO strategy for MOCOP neural solvers.


\section{Weight-Conditioned Neural Solver}

\label{sec4}
 
The architecture of WeCon is schematically depicted in Figure~\ref{fig:WeCon}(a). Its source code is available in the supplementary material. Appendix~\ref{sec3} presents preliminaries on Pareto solutions of MOCOPs and the adopted decomposition method.
\subsection{The Encoder of WeCon}
\label{sec4.1}

Compared with neural solvers for SOCOPs, those tailored for MOCOPs need to incorporate the weight vector $\lambda_s$ associated with each decomposed subproblem in the encoder. However, existing solvers typically only perform a single interaction between instance features and weights in each encoder layer, yielding context that may be insufficiently conditioned on the weights (i.e., low resonance between instance and weight embeddings, see Figure~\ref{fig:embed}).
To better produce more informative weight-conditioned context, we design a novel encoder architecture. Specifically, given a subproblem $({G}=\{v_1, \cdots, v_n \}, \lambda_s)$, the encoder first exploits two Fully Connected Networks (FCNs) to embed the instance features and the weight vector, producing instance embedding  $\bm{H}^0$ and weight embedding $\bm{A}^0$, respectively. Formally,
\begin{equation}
    \bm{H}^0 =  {G}\bm{W}^0 + \bm{b}^0,  \quad 
    \bm{A}^0 = \lambda_s \bm{W}_{\lambda}^0 + \bm{b}_{\lambda}^0,  
\end{equation}
where matrices $\bm{W}^0$, $\bm{b}^0$,$ \bm{W}_{\lambda}^0,$ and $\bm{b}_{\lambda}^0$ are learnable.

 Subsequently, the instance and weight embeddings are jointly updated via $L$ number of layers to produce informative weight-conditioned context, where each layer consists of an MHSA block, two MHA blocks, and our proposed GRF block. Specifically, for the $l$th layer, the instance embeddings $\bm{H}^{l}_1$ ($l\in\{1,2,\cdots L\}$) are first updated via an MHSA block, defined as follows:
\begin{equation}
\bar{\bm{H}}^{l}_1 = \operatorname{RN}\left(\bm{H}^{l-1} + \operatorname{MHSA}\left( \bm{H}^{l-1}, \bm{H}^{l-1}, \bm{H}^{l-1} \right) \right),
\end{equation}
\begin{equation}
\bm{H}^{l}_1 = \operatorname{RN}\left( \bar{\bm{H}}_1^{l} + \operatorname{FF}\left( \bar{\bm{H}}_1^{l} \right) \right),
\end{equation}
\begin{equation}
\footnotesize 
\label{mha}
    \operatorname{MHSA}(\bm{Q},\bm{K},\bm{V})=\left( ||_{m=1}^{M} \operatorname{ATT}_{m}(\bm{Q},\bm{K},\bm{V}) \right) \bm{W}_{o} +\bm{b}_o,
\end{equation}
\begin{equation}
\footnotesize   
\operatorname{ATT}_{m}(\bm{Q},\bm{K},\bm{V}) =\mathrm{S}\left(
\frac{
\left(\bm{Q}\bm{W}_{Q,m}\right)
\left(\bm{K}\bm{W}_{K,m}\right)^{\top}
}{\sqrt{d'}}
\right)\bm{V}\bm{W}_{V,m}.
\end{equation}
where $\operatorname{RN}$, $\operatorname{FF}$, and $\mathrm{S}$ denote the Root Mean Square Layer Normalization \cite{zhang2019root}, Feed Forward Network (FFN), and softmax functions, respectively. The matrices $\bm{W}_{o}$, $\bm{b}_o$,  $\bm{W}_{Q,m}$, $\bm{W}_{K,m}$, and $\bm{W}_{V,m}$ are learnable, where $m$ denotes the $m$th attention head. Symbol $d' = d/M$ denotes the per-head dimension, where $d$ denotes the hidden dimension size. Operators $||$ and $\top$ denote the concatenation and transpose operators, respectively. 
Following \cite{berto2025routefinder}, we adopt SwiGLU \cite{SwishGLU} as the FFN in MHSA and subsequent MHA blocks (MHSA  uses the same source for $Q,K$, and $V$, whereas MHA draws $Q$ and $K/V$ from different sources), which employs SiLU \cite{ELFWING20183} to model the nonlinearity, defined as follows:
\begin{equation}
\footnotesize 
\mathrm{SwiGLU}(\bm{X}) = \bm{X} \odot \sigma(\bm{X}\bm{W}_1 + \bm{b}_1) \otimes \mathrm{SiLU}(\bm{X}\bm{W}_2 + \bm{b}_2),
\end{equation} 
where symbols $\odot$, $\otimes$, and $\sigma$ denote element-wise multiplication, matrix multiplication, and the sigmoid function, respectively. The matrices $\bm{W}_1$, $\bm{b}_1$, $\bm{W}_2$, and $\bm{b} _2$ are learnable. 

Then, we employ two MHA blocks to facilitate interactions between the instance and weight embeddings. Formally,
\begin{equation}
\bar{\bm{A}}^{l} = \operatorname{RN}\left(\bm{A}^{l-1} + \operatorname{MHA}\left( \bm{A}^{l-1}, \bm{H}^{l}_1, \bm{H}^{l}_1 \right) \right),
\end{equation}
\begin{equation}
\bm{A}^{l} = \operatorname{RN}\left( \bar{\bm{A}}^{l} + \operatorname{FF}\left( \bar{\bm{A}}^{l} \right) \right),
\end{equation}
\begin{equation}
\bar{\bm{H}}^{l}_2 = \operatorname{RN}\left(\bm{H}^{l}_1 + \operatorname{MHA}\left( \bm{H}^{l}_1, \bm{A}^{l}, \bm{A}^{l} \right) \right),
\end{equation}
\begin{equation}
\bm{H}^{l}_2 = \operatorname{RN}\left( \bar{\bm{H}}_2^{l} + \operatorname{FF}\left( \bar{\bm{H}}_2^{l} \right) \right).
\end{equation}
By adopting this design, we obtain the weight embedding $\bm{A}^{l}$, which is conditioned on the interactions with the instance embedding. The final instance embedding $\bm{H}^{l}$ is then produced by the proposed  GRF block, defined as follows:
\begin{equation}
 \bm{H}^{l} = \bm{H}_2^{l}  +  g^l \odot(\bm{A}^{l} \bm{W}_3),
\end{equation}
\begin{equation}
\footnotesize 
g^l = \sigma \bigg( \Big(\mathrm{GeLU}\big(([ \bm{H}^{l}_2 || \bm{A}_{\rightarrow n}^{l}])\bm{W}_4+\bm{b}_4 \big) \Big)\bm{W}_5+\bm{b}_5 \bigg), 
\end{equation}
where $\bm{A}_{\rightarrow n}^{l}$ denotes the broadcasted version of $\bm{A}^{l}$ along the node dimension, thus, its shape matches that of $\bm{H}^{l}_2$.
The matrices $\bm{W}_{3}$, $\bm{W}_{4}$, $\bm{b}_{4}$, $\bm{W}_{5}$, and $\bm{b}_{5}$ are learnable. This gated residual design adaptively controls how much weight features are injected into each node embedding.
The architecture of the proposed GRF module is illustrated in Figure~\ref{fig:WeCon}(b).

\subsection{The Decoder of WeCon}
\label{sec4.2}

After obtaining instance and weight embeddings $\bm{H}$ and $\bm{A}$, the decoder autoregressively computes the probability $p_i^t$ of selecting the $i$th node at the $t$th step. To this end, the decoder first produces a vector $\bm{q}_c^t$, then computes attention scores $\alpha^t$ over all candidate nodes to derive $p_i^t$. To produce $\bm{q}_c^t$, prior studies \cite{WE,CNH} rely solely on an MHA layer, whose attention can be dominated by instance embeddings and may therefore dilute the weight signals. To mitigate such dilution, we propose the RF block that further incorporates the weight embedding $\bm{A}$ into the decision process. As discussed in Section~\ref{sec1}, the adoption of RF makes WeCon substantially  differ from existing neural solvers that primarily utilize weights in the encoder. Specifically, WeCon first applies an MHA layer with $M$ attention heads, followed by the RF block. The embedding $\hat{\bm{H}}_1^t$ is computed by the MHA layer as follows:
\begin{equation}
\hat{\bm{H}}_1^t =  \operatorname{MHA}\left(\bm{h}_q^t, [\bm{H}||\bm{A}], [\bm{H}||\bm{A}] \right),
\end{equation}
where $\bm{h}_q^t$ denotes the query embedding. Following the prior studies \cite{WE,POCCO}, we adopt problem-specific query embeddings for different MOCOP variants. 
Subsequently, to enable weight-conditioned decisions, we feed $\hat{\bm{H}}_1^t$ into the RF block.  Formally,  
\begin{equation}
\footnotesize 
 \bm{q}_c^t =\hat{\bm{H}}_1^t +\Big( \big(\mathrm{ReLU}([\hat{\bm{H}}_1^t||\bm{A}_{\rightarrow n}]\bm{W}_6+\bm{b}_6)\big)\bm{W}_7 +\bm{b}_7 \Big  
),
\end{equation}
where $\bm{W}_{6}$, $\bm{b}_{6}$, $\bm{W}_{7}$, and $\bm{b}_{7}$ are learnable.  Figure~\ref{fig:WeCon}(c) illustrates RF's architecture. By adopting this design, WeCon mitigates weight-signal dilution, enabling weight-conditioned decisions (see Section~\ref{sec5.1}). 
Moreover, the effect of RF can be analyzed from a theoretical intuition. With {RF}, $q_c^t = \operatorname{MHA}(\cdot)+ \operatorname{RF}(\operatorname{MHA}(\cdot) || A_n)$. Accordingly, we have 
\begin{equation}
\label{eq15}
    \frac{\partial q_c^t}{\partial A} = \frac{\partial \operatorname{MHA}}{\partial A} + \frac{\partial \operatorname{RF}}{\partial \operatorname{MHA}} \cdot \frac{\partial \operatorname{MHA}}{\partial A} + \frac{\partial {\operatorname{RF}}}{\partial A_n} \cdot \frac{\partial A_n}{\partial A}.
\end{equation}
Compared to decoders without RF, the last term of Eq.~(\ref{eq15}) introduces an additional direct gradient path from $A$ to $q_c^t$. Because $A_n$ is directly broadcast from $A$,   $\frac{\partial A_n}{\partial A}$ is non-zero. Under a mild non-degeneracy assumption, $\left\|\frac{\partial \operatorname{RF}}{\partial A_n}\right\|_2 \ge u > 0$, where $u$ is a positive constant, this path remains locally non-vanishing. Thus, RF helps mitigate weight-signal dilution during decoding.


Finally, the probability $p_i^t$ is computed as follows:
\begin{equation}
\label{pi}
    p_i^t = p_{\theta}\left(\pi_t=i \mid {G}, \lambda, \pi_{1:t-1}\right)
= \frac{e^{\alpha_i^t}}{\sum_{j} e^{\alpha_j^t}},
\end{equation}
\begin{equation}
\alpha_i^t =
\begin{cases}
-\infty , & \text{if the $i$th node is masked},
\\
C \cdot \tanh (\frac{\bm{q}_c^{\top}  \bm{h}_i }{\sqrt{d}}), &
\text{otherwise,} 
\end{cases}
\end{equation}
where $\theta$ denotes all learnable model parameters and $C$ is a predefined clipping constant following prior studies \cite{WE,CNH}. 

Notably, RF is a plug-and-play module that can be effortlessly integrated into various decoder architectures. To demonstrate its generality, we propose WeCon-CCO, which inserts RF between the two components of the decoder in POCCO-W \cite{POCCO}, i.e., the MHA layer and the CCO module. Please see Appendix~\ref{cco} for the detailed architecture of WeCon-CCO. With this enhanced decoder, WeCon-CCO achieves the best overall HV performance across different MOCOPs, at the cost of requiring significantly longer inference time than WeCon (see Section~\ref{sec5.1}).

\subsection{Effective Preference Optimization (EPO)}
\label{sec4.3}



To construct informative preference pairs as training signals for solver optimization, we propose the EPO strategy. Instead of sampling $r$ solutions purely at random for each instance, EPO performs a guided sampling strategy to obtain $\lceil \frac{r}{c} \rceil$ number of high-quality solutions (see Eq.~(\ref{sample})), while the others ($r-\lceil \frac{r}{c} \rceil $) are sampled randomly. Here, $c>1$ controls the fraction of guided samples. Specifically, given the probabilities $p_i^t$ (see Eq.~(\ref{pi})) over all unmasked (feasible) nodes at the $t$th step, guided sampling constrains the next-node decision to top-$k$ nodes with the highest probabilities. Formally, in the $t$th step, the candidate node set is defined as follows:
\begin{equation}
\label{sample}
\mathcal{T}^t = \{i \mid i \in \mathop{\operatorname{arg\,top(\textit{k})}}\limits_{j \in G^t} p_j^t \},
\end{equation}
where $G^t$ denotes the set of feasible nodes at the $t$th step. If the number of remaining feasible nodes is less than $k$,  i.e., $|G^t|<k$, we simply set $\mathcal{T}^t=G^t$. Hence, EPO samples a sufficient number of high-quality solutions, yielding more informative preference pairs in which the better solution $\pi_w$ substantially outperforms the worse solution $\pi_l$ in terms of solution quality (see Table~\ref{ablation}). Subsequently, these guided and randomly sampled solutions are jointly used to construct a number of $\frac{r(r-1)}{2}$ preference pairs via pairwise comparisons.

\begin{table*}[!t]
  \centering
  \caption{Performance comparison of different methods on Bi-TSP20, Bi-TSP50, and Bi-TSP100}
  \label{tab:bitsp}
  \vspace{-0.3cm}
  \begin{tabular}{l ccc ccc ccc}
    \toprule
    \multirow{2}{*}{Method} 
    & \multicolumn{3}{c}{Bi-TSP20}
    & \multicolumn{3}{c}{Bi-TSP50}
    & \multicolumn{3}{c}{Bi-TSP100} \\
    \cmidrule(lr){2-4}\cmidrule(lr){5-7}\cmidrule(lr){8-10}
    & HV$\uparrow$ & Gap$\downarrow$ & Time$\downarrow$
    & HV$\uparrow$ & Gap$\downarrow$ & Time$\downarrow$
    & HV$\uparrow$ & Gap$\downarrow$ & Time$\downarrow$ \\
    \midrule
    WS-LKH & 0.6270 & 0.00\% & 10m & 0.6415 & 0.03\% & 1.8h & \textbf{0.7090} & \textbf{-0.16\%} & 6h \\
    NSGA-II (TEVC'02) & 0.6258 & 0.19\% & 6.0h & 0.6120 & 4.63\% & 6.1h & 0.6692 & 5.47\% & 6.9h \\
    MOGLS (EJOR'02) & \textbf{0.6279} & \textbf{-0.14\%} & 1.6h & 0.6330 & 1.36\% & 3.7h & 0.6854 & 3.18\% & 11h \\
    MOEA/D (TEVC'07) & 0.6241 & 0.46\% & 1.7h & 0.6316 & 1.57\% & 1.8h & 0.6899 & 2.54\% & 2.2h \\
    PPLS/D-C (TCYB'24) & 0.6256 & 0.22\% & 26m & 0.6282 & 2.10\% & 2.8h & 0.6844 & 3.32\% & 11h \\
    \midrule
    DRL-MOA (TCYB'21) & 0.6257 & 0.21\% & 6s & 0.6360 & 0.89\% & 9s & 0.6970 & 1.54\% & 16s \\
    MDRL (TNNLS'23) & 0.6271 & -0.02\% & 5s & 0.6364 & 0.83\% & 8s & 0.6969 & 1.55\% & 14s \\
    \rowcolor{lightgray} MDRL-A (TNNLS'23) & 0.6271 & -0.02\% & 47s & 0.6408 & 0.14\% & 1.8m & 0.7022 & 0.81\% & 5.4m \\
    EMNH (NeurIPS'23) & 0.6271 & -0.02\% & 5s & 0.6364 & 0.83\% & 8s & 0.6969 & 1.55\% & 15s \\
    \rowcolor{lightgray} EMNH-A (NeurIPS'23) & 0.6271 & -0.02\% & 46s & 0.6408 & 0.14\% & 1.8m & 0.7023 & 0.79\% & 5.4m \\
    \midrule
    PMOCO (ICLR'22) & 0.6256 & 0.22\% & 2s & 0.6354 & 0.98\% & 4s & 0.6969 & 1.55\% & 11s \\
    \rowcolor{lightgray} PMOCO-A (ICLR'22) & 0.6270 & 0.00\% & 36s & 0.6395 & 0.34\% & 1.9m & 0.7037 & 0.59\% & 9m \\
    CNH (TNNLS'25) & 0.6270 & 0.00\% & 3s & 0.6387 & 0.47\% & 6s & 0.7019 & 0.85\% & 15s \\
    \rowcolor{lightgray} CNH-A (TNNLS'25) & 0.6271 & -0.02\% & 39s & 0.6410 & 0.11\% & 2.9m & 0.7054 & 0.35\% & 12m \\
    WE-CA (ICLR'25) & 0.6271 & -0.02\% & 2s & 0.6392 & 0.39\% & 4s & 0.7034 & 0.64\% & 11s \\
    \rowcolor{lightgray} WE-CA-A (ICLR'25) & 0.6271 & -0.02\% & 35s & 0.6413 & 0.06\% & 2.3m & 0.7066 & 0.18\% & 10m \\
    PA-MoE-W (OpenReview'25) & 0.6271 & -0.02\% & 6s & 0.6396 & 0.33\% & 13s & 0.7042 & 0.52\% & 25s \\
    \rowcolor{lightgray} PA-MoE-W-A (OpenReview'25) & 0.6271 & -0.02\% & 1.6m & \textbf{0.6425} & \textbf{-0.12\%} & 4.6m & 0.7070 & 0.13\% & 19m \\
        GF-MOCA (OpenReview'25) &  0.6272 & -0.03\% & 7s & 0.6393 & 0.37\% & 14s & 0.7037 & 0.59\% & 27s \\
    \rowcolor{lightgray} GF-MOCA-A (OpenReview'25) & 0.6272 & -0.03\% & 2.2m & 0.6414 & 0.05\% &  5.0m &0.7065 & 0.20\% &14m \\
    POCCO-W (NeurIPS'25) & \underline{0.6275} & \underline{-0.08\%} & 6s & 0.6411 & 0.09\% & 10s & 0.7054 & 0.35\% & 27s \\
    \rowcolor{lightgray} POCCO-W-A (NeurIPS'25) & 0.6270 & 0.00\% & 1.5m & \underline{0.6418} & \underline{-0.02\%} & 5.6m & 0.7077 & 0.03\% & 21m \\
    \midrule
    \textbf{WeCon} (ours) & 0.6271 & -0.02\% & 3s & 0.6407 & 0.16\% & 6s & 0.7056 & 0.32\% & 15s \\
    \rowcolor{lightgray} \textbf{WeCon-A} (ours) & 0.6270 & 0.00\% & 40s & 0.6415 & 0.03\% & 3m & 0.7077 & 0.03\% & 12m \\
    \textbf{WeCon-CCO} (ours) & 0.6273 & -0.05\% & 5s & 0.6412 & 0.08\% & 11s & 0.7061 & 0.25\% & 30s \\
    \rowcolor{lightgray} \textbf{WeCon-CCO-A} (ours) & 0.6270 & 0.00\% & 1.4m & 0.6417 & 0.00\% & 6.2m & \underline{0.7079} & \underline{0.00\%} & 23m \\
    \bottomrule
  \end{tabular}
  \vspace{-0.2cm}
\end{table*}

To learn from preference pairs, EPO treats the average log likelihood of a solution as an implicit reward $f_\theta(\pi\mid{G},\lambda_s)$, thereby linking the solution preferences to their policy probabilities, defined as follows:
\begin{equation}
\label{eq:simpo_reward}
\footnotesize 
    f_{\theta} =\frac{\log p_{\theta}(\pi\mid{G}, \lambda_s)}{|\pi|} = \frac{\sum_{t=1}^{|\pi|}\log\, p_\theta(\pi_t\mid \pi_{<t}, {G},\lambda_s)}{|\pi|},
\end{equation}
where $p_{\theta}(\pi\mid{G}, \lambda_s)$ denotes the probability of generating solution $\pi$ and  $|\pi|$ denotes the sequence length of $\pi$ (used to normalize the implicit reward). Then, we define $g_\theta(\cdot)$ to map reward difference $f_\theta(\pi_w\mid{G},\lambda_s) -f_\theta(\pi_l\mid{G},\lambda_s)$ to a preference probability as follows:
\begin{equation}
\footnotesize 
g_{\theta}\!\bigl(\pi_{w}\lessdot\pi_{l}\mid{G},\lambda_s \bigr)=\sigma \Big( \beta \big( f_{\theta}(\pi_{w}\mid{G},\lambda_s) - f_{\theta}(\pi_{l}\mid{G},\lambda_s) \big) \Big),
\end{equation}
where $\sigma$ denotes the sigmoid function and $\beta$ denotes a predefined temperature parameter that controls the sharpness of the preference comparison. 
Finally, WeCon is trained by maximizing the log likelihood of $g_\theta(\cdot)$ and the loss function defined as follows:
\begin{equation}
\label{eq:loss}
\footnotesize 
\begin{aligned}
& \mathcal{L}(\theta\mid p_\theta, {G},\lambda_s, \pi_w, \pi_l) = 
 \\ &-y\log\sigma \bigl( \beta [\frac{\log\, p_\theta(\pi_w\mid{G},\lambda_s)}{|\pi_w|}-\frac{\log\, p_\theta(\pi_l\mid{G},\lambda_s)}{|\pi_l|}\bigr]),
\end{aligned}
\end{equation}
where  $y$ denotes a binary preference label that $y = 1$ if $\pi_{w} \lessdot \pi_{l}$, and $y = 0$ otherwise.
The pseudocode of EPO is presented in Appendix~\ref{alg:epo}.

\section{Experimental Results}
\label{sec5}

In this section, to assess the performance of the proposed WeCon and \mbox{WeCon-CCO}, we compare them against fourteen methods on synthetic Bi-objective Traveling Salesman Problem (Bi-TSP) instances across five scales and real-world Bi-TSP instances. We further provide visual evidence to demonstrate the effectiveness of WeCon’s encoder. Finally, we conduct ablation studies to validate the contributions of all proposed components.
Detailed experimental setups are provided in Appendix~\ref{app:setup}. Appendix~\ref{otherMOCOPS} reports the performance of different methods on the Bi-objective Capacitated Vehicle Routing Problem (Bi-CVRP), the Bi-objective Knapsack Problem (Bi-KP), and the Tri-objective Traveling Salesman Problem (Tri-TSP). Appendix~\ref{sec5.2} shows the performance of different methods on large-scale Bi-TSP and Bi-KP instances.


\subsection{Performance Comparisons}
\label{sec5.1}
\begin{table}[!t]
  \centering
  \caption{Performance comparison of different methods on Bi-TSP150 and Bi-TSP200}
  \vspace{-0.3cm}
  \setlength{\tabcolsep}{0.5mm}
  \label{tab:bitsp-med}
  \begin{tabular}{l ccc ccc}
    \toprule
    \multirow{2}{*}{Method} 
    & \multicolumn{3}{c}{Bi-TSP150}
    & \multicolumn{3}{c}{Bi-TSP200} \\
    \cmidrule(lr){2-4}\cmidrule(lr){5-7}
    & HV$\uparrow$ & Gap$\downarrow$ & Time
    & HV$\uparrow$ & Gap$\downarrow$ & Time \\
    \midrule
    WS-LKH & \textbf{0.7149} & \textbf{-1.17\%} & 13h & \textbf{0.7490} & \textbf{-1.15\%} & 22h \\
    NSGA-II  & 0.6659 & 5.76\% & 6.8h & 0.7045 & 4.86\% & 6.9h \\
    MOGLS & 0.6768 & 4.22\% & 22h & 0.7114 & 3.93\% & 38h \\
    MOEA/D  & 0.6809 & 3.64\% & 2.4h & 0.7139 & 3.59\% & 2.7h \\
    PPLS/D-C  & 0.6784 & 3.99\% & 21h & 0.7106 & 4.04\% & 32h \\
    \midrule
    DRL-MOA  & 0.6901 & 2.34\% & 36s & 0.7219 & 2.51\% & 1.2m \\
    MDRL & 0.6922 & 2.04\% & 36s & 0.7251 & 2.08\% & 1.1m \\
    \rowcolor{lightgray} MDRL-A  & 0.6976 & 1.27\% & 37m & 0.7299 & 1.43\% & 1.1h \\
    EMNH  & 0.6930 & 1.92\% & 37s & 0.7260 & 1.96\% & 1.1m \\
    \rowcolor{lightgray} EMNH-A  & 0.6983 & 1.17\% & 39m & 0.7307 & 1.32\% & 1.1h \\
    \midrule
    PMOCO  & 0.6910 & 2.21\% & 31s & 0.7231 & 2.35\% & 1.0m \\
    \rowcolor{lightgray} PMOCO-A  & 0.6967 & 1.40\% & 32m & 0.7283 & 1.65\% & 1.1h \\
    CNH & 0.6985 & 1.15\% & 35s & 0.7292 & 1.53\% & 1.1m \\
    \rowcolor{lightgray} CNH-A  & 0.7025 & 0.58\% & 34m & 0.7343 & 0.84\% & 1.2h \\
    WE-CA  & 0.7008 & 0.82\% & 30s & 0.7346 & 0.80\% & 1.0m \\
    \rowcolor{lightgray} WE-CA-A  & 0.7044 & 0.31\% & 31m & 0.7381 & 0.32\% & 1.0h \\
    PA-MoE-W  & 0.7019 & 0.67\% & 59s & 0.7360 & 0.61\% & 1.7m \\
    \rowcolor{lightgray} PA-MoE-W-A & 0.7052 & 0.20\% & 55m & 0.7391 & 0.19\% & 1.9h \\
        GF-MOCA  & 0.7008 & 0.82\% & 1.4m & 0.7343 & 0.84\% & 2.4m \\
     POCCO-W  & 0.7029 & 0.52\% & 1.0m & 0.7362 & 0.58\% & 1.8m \\
    \rowcolor{lightgray} POCCO-W-A  & 0.7062 & 0.06\% & 59m & 0.7399 & 0.08\% & 2.1h \\
    \midrule
    \textbf{WeCon}  & 0.7035 & 0.44\% & 36s & 0.7374 & 0.42\% & 1.2m \\
    \rowcolor{lightgray} \textbf{WeCon-A} & 0.7063 & 0.04\% & 36m & 0.7402 & 0.04\% & 1.2h \\
    \textbf{WeCon-CCO} & 0.7040 & 0.37\% & 1.1m & 0.7379 & 0.35\% & 1.9m \\
    \rowcolor{lightgray} \textbf{WeCon-CCO-A} & \underline{0.7066} & \underline{0.00\%} & 58m & \underline{0.7405} & \underline{0.00\%} & 1.8h \\
    \bottomrule
  \end{tabular}
   \vspace{-0.3cm}
\end{table}
 In Tables~\ref{tab:bitsp} and \ref{tab:bitsp-med}, we report the results of different methods on small- and medium-scale Bi-TSP, respectively.  
The baseline models include five conventional heuristics, namely WS-LKH, MOEA/D \cite{zhang_MOEA_2007}, NSGA-II \cite{deb2002fast}, MOGLS \cite{jaszkiewicz2002genetic}, and PPLS/D-C \cite{shi2022improving}, and nine neural solvers, namely DRL-MOA \cite{Li_deep_2021}, MDRL \cite{Zhang_meta_2022}, EMNH \cite{chen2023efficient}, PMOCO \cite{lin_pareto_2022}, CNH \cite{CNH}, WE-CA \cite{WE}, PA-MoE-W \cite{iclr2026}, GF-MOCA \cite{GF-MOCA}, and POCCO-W \cite{POCCO}. Among these models, DRL-MOA, MDRL, and EMNH are multi-model neural solvers that train distinct models for different subproblems, whereas PMOCO, CNH, WE-CA, PA-MoE-W, GF-MOCA, and \mbox{POCCO-W}, as well as our WeCon and WeCon-CCO are single-model solvers that use a unified model to handle all subproblems. WS-LKH is the SOTA heuristic widely used as a baseline \cite{ POCCO}. POCCO-W is the SOTA neural solver for MOCOPs. Notably, CNH, WE-CA, PA-MoE-W, GF-MOCA, POCCO-W, WeCon, and WeCon-CCO are trained on multiple problem scales $n \in \{20, 21, \cdots, 100\}$ to obtain a unified model applicable across individual scales.  The resulting models are then directly evaluated on test datasets of varying sizes without additional fine-tuning for specific scales. Following the prior arts in the field, we evaluate all methods using three metrics, namely average HyperVolume (HV) \cite{while2006faster}, average gap, and total runtime per test dataset. HV is a widely used indicator in MOCOPs that reflects both the convergence and diversity of the obtained solution set; higher HV indicates better performance. For consistency, HV is normalized to $[0,1]$ using the same reference and ideal points for all methods (see Appendix~\ref{app:setup}). The gap is defined as the relative difference in HV between the corresponding method and WeCon-CCO-A. Methods with the “-A” suffix apply instance augmentation \cite{lin_pareto_2022} to further improve performance and are highlighted with a gray background in the tables (see Appendix~\ref{app:setup} for augmentation details). Moreover, following \cite{WE,POCCO}, we conduct the Wilcoxon rank-sum test at the
1\% significance level. In each comparison, the best result is highlighted in bold only when it shows a statistically
significant improvement over the other methods, while the second-best result is underlined.
 


 As shown in Tables~\ref{tab:bitsp} and \ref{tab:bitsp-med}, our WeCon-A achieves HV scores comparable to POCCO-W-A, with relative Gap differences of 0.00\%, -0.05\%, 0.00\%, +0.02\%, and +0.04\% across Bi-TSP20, Bi-TSP50, Bi-TSP100, Bi-TSP150, and Bi-TSP200, respectively, while requiring approximately 55\%, 46\%, 42\%, 38\%, and  43\% less inference time on the corresponding datasets. 
This speedup mainly stems from architectural differences. Specifically, \mbox{POCCO-W} employs the MoE-based CCO module, whose gating-and-routing mechanism assigns subproblems to different experts, incurring additional dispatch/aggregation overhead. This higher runtime may limit its practicality in time-sensitive multi-objective applications, such as real-time urban traffic signal control, where conflicting objectives, including traffic efficiency and safety,  must be jointly optimized and solutions must be recomputed within seconds as traffic conditions evolve rapidly~\cite{10974678}. In contrast, our RF module avoids such routing overhead, delivering faster inference while maintaining high solution quality and diversity by mitigating weight-signal dilution during decoding. 
More importantly, without any data augmentation, WeCon achieves significantly higher HV scores than POCCO-W, reducing the Gap by 0.08\% and 0.16\% on  Bi-TSP150 and Bi-TSP200, respectively. This advantage is mainly attributable to our encoder design, which provides more informative weight-conditioned context. Beyond Bi-TSP,
WeCon-CCO achieves the best overall performance across all MOCOP variants and scales (see Appendix~\ref{otherMOCOPS}), albeit at the cost of increased inference time relative to WeCon (on average, $1.7\times$). Its performance advantage becomes more pronounced as the problem size increases, indicating a stronger ability to explore the solution space and approximate high-quality Pareto fronts across scales (see Appendix~\ref{sec5.2} for results on large-scale Bi-TSP and Bi-KP instances). 
Notably, the performance of WeCon and WeCon-CCO stems from our carefully designed encoder and decoder architectures rather than merely from increased model size. Appendix~\ref{size} provides comparisons with enlarged WE-CA and POCCO-W models.

\begin{table}[!t]
  \centering
  \caption{Performance comparison of different methods on real-world Bi-TSP instances}
      \vspace{-0.3cm}
    \setlength{\tabcolsep}{0.5mm}
  \label{tab:kroab}
    \begin{tabular}{lcc cc cc}
      \toprule
     \multirow{2}{*}{Method} & \multicolumn{2}{c}{KroAB100}
      & \multicolumn{2}{c}{KroAB150}
      & \multicolumn{2}{c}{KroAB200} \\
      \cmidrule(lr){2-3}\cmidrule(lr){4-5}\cmidrule(lr){6-7}
          & HV$\uparrow$      & Time$\downarrow$   & HV$\uparrow$           & Time$\downarrow$   & HV$\uparrow$        & Time$\downarrow$    \\
      \midrule
       WS-LKH    & \textbf{0.7022} & 2.3m   & \textbf{0.7017}  & 4.0m   & \textbf{0.7430} & 5.6m   \\
    MOEA/D   & 0.6836  & 5.8m   & 0.6710 &   7.1m   & 0.7106   & 7.3m   \\
    NSGA-II  & 0.6676 &   7.0m   & 0.6552 &   7.9m   & 0.7011 &   8.4m   \\
    MOGLS    & 0.6817 &   52m    & 0.6671 &  1.3h   & 0.7083 &   1.6h   \\
    PPLS/D-C& 0.6785 &   38m    & 0.6659 &   1.4h   & 0.7100 &   3.8h   \\
    \midrule
    DRL-MOA   & 0.6903 &   10s    & 0.6794 &   12s    & 0.7185 &   18s    \\
    MDRL     & 0.6881 &   9s     & 0.6831 &   11s    & 0.7209 &   16s    \\
 \rowcolor{lightgray}   MDRL-A    & 0.6950 &   10s    & 0.6890 &   16s    & 0.7261 &   25s    \\
    EMNH      & 0.6900 &   9s     & 0.6832 &  11s    & 0.7217 &   16s    \\
 \rowcolor{lightgray}   EMNH-A  & 0.6958 &   10s    & 0.6892 &  16s    & 0.7270 &   25s    \\
    \midrule
    PMOCO    & 0.6878 &   7s     & 0.6819 &  10s    & 0.7193 &   15s    \\
\rowcolor{lightgray}    PMOCO-A & 0.6937 &   11s    & 0.6886 &   15s    & 0.7251 &   24s    \\
    CNH  & 0.6947 &   8s    & 0.6892 &  13s    & 0.7250 &   18s    \\
 \rowcolor{lightgray}   CNH-A   & 0.6980 &   12s    & 0.6938 &   16s    & 0.7303 &  25s    \\
    WE-CA    & 0.6948 &   7s     & 0.6924 &   12s    & 0.7317 &  16s    \\
 \rowcolor{lightgray}   WE-CA-A  & 0.6992 &   10s    & 0.6958 &   15s    & 0.7347 &  24s    \\
    PA-MoE-W  &0.6965 & 19s & 0.6926&29s  &0.7325 &38s \\
\rowcolor{lightgray}    PA-MoE-W-A  &0.6998 &22s &0.6969 &33s &0.7361 &48s\\

  GF-MOCA  &0.6955 & 17s & 0.6915  &22s  &0.7306 &31s \\
\rowcolor{lightgray}    GF-MOCA-A  &0.6993 &43s &0.6960 &2.0m &0.7339 &5m\\

    POCCO-W   & 0.6978 &  20s    & 0.6938 &  31s    & 0.7331 &   41s    \\
\rowcolor{lightgray}    POCCO-W-A  & \underline{0.7006} &  22s   & 0.6971   & 34s     &0.7360 & 50s    \\
    \midrule
    WeCon   & 0.6974 &  8s & 0.6946 &   13s    &  0.7337&  17s    \\
\rowcolor{lightgray}    WeCon-A  & 0.7000 &  10s   &  0.6975 &  16s     &0.7369 & 26s    \\
    WeCon-CCO   & 0.6991 &  19s   & 0.6948 &   30s    &  0.7342&  39s    \\
\rowcolor{lightgray}    WeCon-CCO-A  & \underline{0.7006} &  23s   &  \underline{0.6981}&  34s     &\underline{0.7375} &  50s    \\
      \bottomrule
    \end{tabular}
\end{table}

Moreover, we evaluate the cross-distribution generalization of WeCon and WeCon-CCO. Following the prior arts \cite{WE,POCCO}, we select three widely benchmarked real-world instances adapted from TSPLIB \cite{tsplib}, namely KroAB100, KroAB150, and KroAB200. These instances exhibit underlying distributions that differ substantially from those of synthetic datasets and are difficult to characterize. As shown in Table~\ref{tab:kroab}, WeCon surpasses POCCO-W on KroAB150 and KroAB200. WeCon-CCO achieves the best overall performance, demonstrating strong cross-distribution generalization. In addition, Appendix~\ref{sec5.3} visualizes the Pareto fronts produced by different methods on these three instances. 

\subsection{Visualization Analysis}
\begin{figure}[!t]
\centering
\subfloat{
\includegraphics[width=0.24\textwidth]{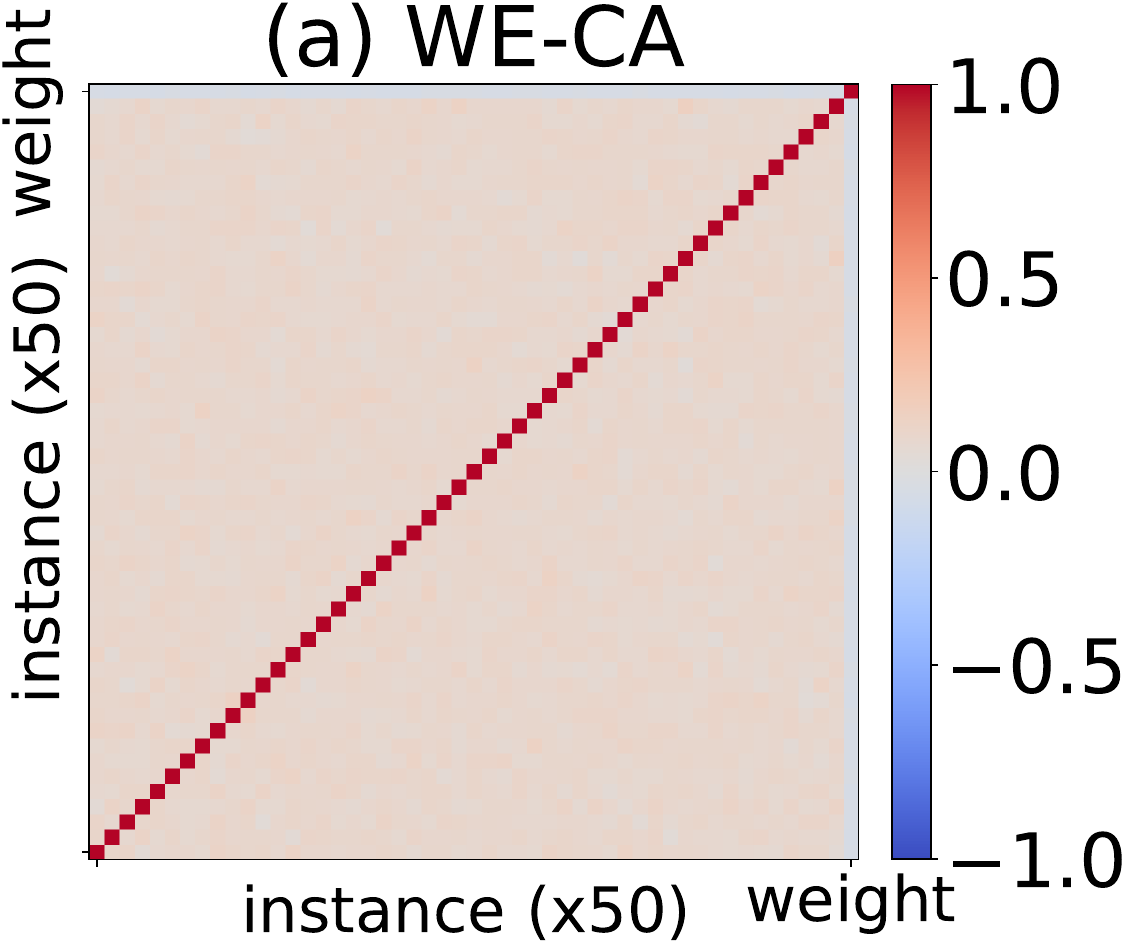}
}
\subfloat{
\includegraphics[width=0.24\textwidth]{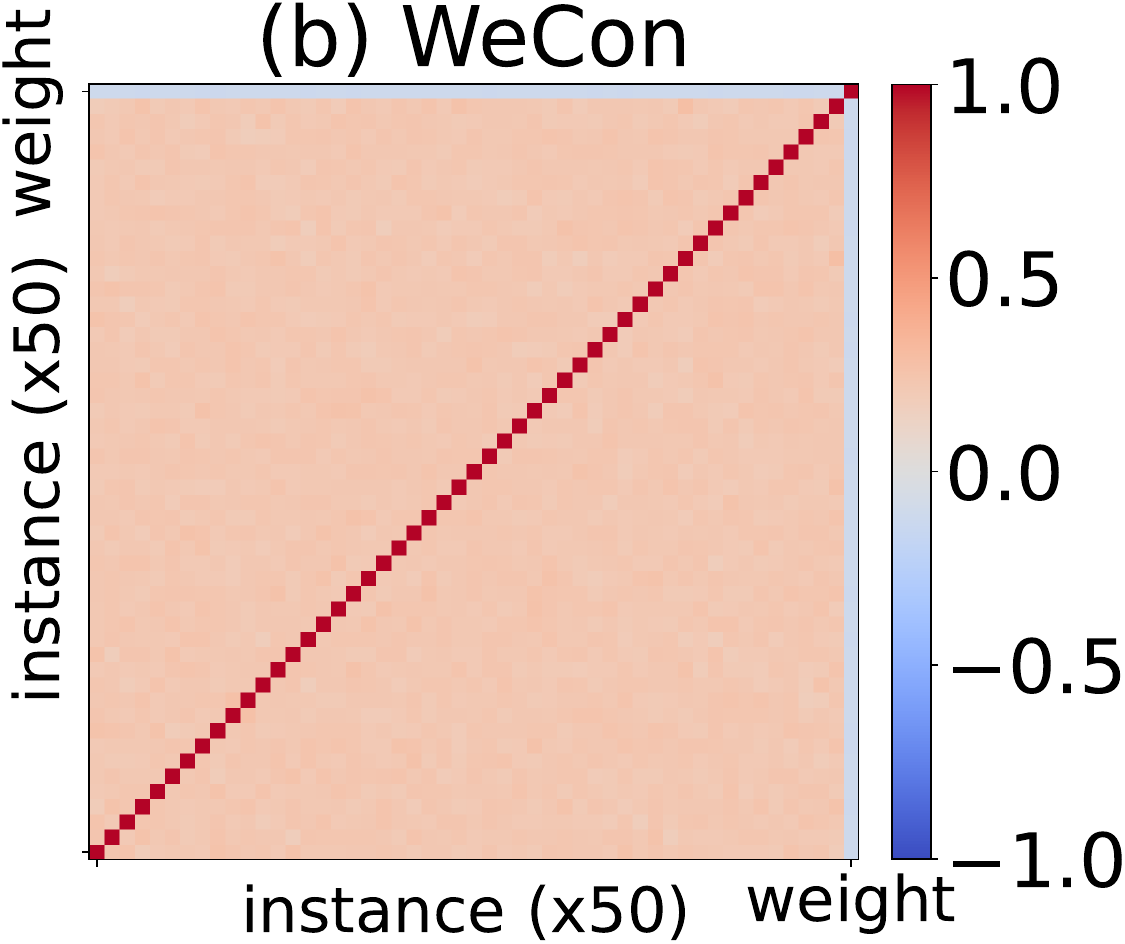}
}
\vspace{-0.2cm}
\caption{Similarity between weight \& instance embeddings.}
\label{fig:embed}
\vspace{-0.3cm}
\end{figure}

In Figure~\ref{fig:embed}, we visualize the cosine similarity between the weight embeddings and instance embeddings produced by WE-CA and WeCon, respectively.
Compared with WE-CA, WeCon exhibits warmer colors in the off-diagonal regions among instance elements, i.e., the 50 nodes,  while showing cooler tones along the edges between the weight and instance embeddings.  These results suggest that WeCon achieves higher representation resonance and learns substantially more discriminative weight and instance embeddings. 
In addition, Appendix~\ref{appendixH} provides visualization results to validate that the proposed RF module can mitigate weight-signal dilution during decoding.



\subsection{Ablation Studies}
\label{sec5.4}
\begin{table}[!t]
    \centering
    \caption{Ablation study results on different design choices}
    \vspace{-0.3cm}
\setlength{\tabcolsep}{0.2mm}
    \label{ablation}
    \begin{tabular}{l cc cc cc cc}
        \toprule
      \multirow{2}{*}{Method}  
      & \multicolumn{4}{c}{Bi-TSP100}
      & \multicolumn{4}{c}{Bi-TSP150}  \\
      \cmidrule(lr){2-5}\cmidrule(lr){6-9} 
         & HV$\uparrow$ & Time   & HV$\uparrow$ & Time & HV$\uparrow$ & Time & HV$\uparrow$ & Time \\ \toprule
        w/o Enc.    &0.7050 &13s &\cellcolor{lightgray}  0.7073 &\cellcolor{lightgray} 11m & 0.7028  &35s &\cellcolor{lightgray} 0.7058 &\cellcolor{lightgray} 35m\\ \midrule
        w/o GRF    & 0.7053& 15s &\cellcolor{lightgray} 0.7075  & \cellcolor{lightgray} 11m & 0.7032   &   36s&\cellcolor{lightgray}  0.7060  &\cellcolor{lightgray}  36m\\ \midrule
        w/o BC    & 0.7046 & 14s &\cellcolor{lightgray} 0.7070   & \cellcolor{lightgray}  12m &  0.7019  &  36s &\cellcolor{lightgray}   0.7051&\cellcolor{lightgray} 36m \\ \midrule
        w/o RF   &0.7045 & 12s &\cellcolor{lightgray} 0.7070  & \cellcolor{lightgray} 9.8m &  0.7020 & 30s  &\cellcolor{lightgray} 0.7051&\cellcolor{lightgray}30m   \\ \midrule
        w/ RL    & 0.7053  & 14s &\cellcolor{lightgray}0.7074   &\cellcolor{lightgray}12m &  0.7030  & 36s  &\cellcolor{lightgray} 0.7059 &\cellcolor{lightgray}36m \\ \midrule
        w/ PO    & 0.7052  & 15s & \cellcolor{lightgray}0.7075  &\cellcolor{lightgray}12m &0.7030  &  36s &\cellcolor{lightgray} 0.7060   &\cellcolor{lightgray}36m  \\ \midrule
        w/ BOPO   & 0.7025  & 15s &\cellcolor{lightgray}  0.7058 &\cellcolor{lightgray} 12m &0.6991   & 36s& \cellcolor{lightgray} 0.7030  & \cellcolor{lightgray} 36m\\ \midrule
        WeCon  & \textbf{0.7056}   &15s   & \cellcolor{lightgray} \textbf{0.7077}  &  \cellcolor{lightgray}12m &  \textbf{0.7035}  &  36s  &\cellcolor{lightgray} \textbf{0.7063}  &  \cellcolor{lightgray} 36m\\ \bottomrule
    \end{tabular}
    {
    \\ \textit{Note:}  Results with augmentation are highlighted in gray.}
        \vspace{-0.3cm}
\end{table}

In this subsection, we conduct ablation studies to validate the effectiveness of WeCon’s encoder and decoder designs, as well as the proposed EPO strategy. Specifically, w/o Enc. refers to the setting that replaces the proposed encoder with the encoder of \cite{WE}; w/o GRF refers to the setting that the encoder only comprises an MHSA block and two MHA blocks; w/o BC refers to the setting that the encoder only comprises an MHSA block and GRF block;  w/o RF refers to the setting that the decoder only comprises a single MHA layer to compute the vector $\bm{q}_c^t$; w/ RL, w/ PO, and w/ BOPO refer to the settings that replace the proposed EPO strategy with REINFORCE, PO \cite{PO}, and BOPO \cite{BOPO}, respectively. As shown in Table~\ref{ablation}, it is not surprising that removing any component in WeCon always
degrades performance. Moreover, the proposed EPO strategy outperforms all the other widely adopted training strategies. This improvement stems from EPO’s ability to sample sufficiently high-quality solutions, thereby constructing more informative preference pairs and improving training effectiveness. Overall, these results validate the contributions of all proposed components.

Finally, Appendix~\ref{kc} shows that WeCon is not sensitive to the choices of $k$ and $c$. Appendix~\ref{app:ws} further validates the rationale for adopting WS as the decomposition technique. Appendix~\ref{app:pre} further demonstrates the effectiveness and generality of RF through preliminary experiments.


\section{Conclusion}
In this study, we positively answer the proposed research question with ample experimental results as supporting evidence.
Specifically, we propose WeCon, which comprises a carefully designed encoder to produce informative weight-conditioned context and our proposed decoder to enable
weight-conditioned decisions. Moreover, we introduce the EPO strategy for WeCon, which constructs more informative preference pairs and thereby improves training effectiveness. The experimental results show that i) WeCon achieves HV scores comparable to the SOTA solver POCCO-W while requiring only 60\% of its inference time and ii) WeCon-CCO achieves the best overall performance across all evaluated settings while sacrificing runtime efficiency. Ablation studies validate the effectiveness of our encoder and decoder designs as well as EPO strategy. Currently, WeCon and WeCon-CCO are trained across multiple problem scales to obtain a unified model applicable to all scales. This design can degrade performance on test instances at a particular scale compared with models trained exclusively on that scale. Going forward, we believe further improvement is achievable by adopting Curriculum Learning to progressively train the solver, thereby further enhancing generalization across problem scales.

\bibliography{aaai2027}

\clearpage
\clearpage
\appendix

\section{Preliminaries}
In this section, we introduce Pareto-optimal solutions of MOCOPs and the adopted decomposition method.
\label{sec3}

\noindent\textbf{Pareto-optimal Solutions of MOCOPs}: A MOCOP instance ${G}=\{v_1,\cdots, v_n\}$ can be formulated as $\min_{\pi\in \mathcal{X}}$ $ F(\pi) = (F^{1}(\pi), \dots, F^{\kappa}(\pi))$, where $F$ denotes an objective vector with a number of $\kappa$ objective functions, $\pi$ denotes a feasible solution, and $\mathcal{X}$ denotes the feasible solution space.
Owing to inherent conflicts among objectives, typically there is no single solution that is optimal for all objectives. Therefore, we seek Pareto-optimal solutions that leverage performance among objectives, defined as follows:

\noindent\textit{Pareto Dominance}: A solution $\pi \in \mathcal{X}$ is said to dominate another solution $\pi' \in \mathcal{X}$ (denoted as $\pi \prec \pi'$) iff $F^{i}(\pi) \leq F^{i}(\pi')$,  $\forall i \in \left\{1, \cdots, \kappa\right\}$ and $F(\pi) \neq F(\pi')$.

\noindent\textit{Pareto Optimality:} A solution $\pi^* \in \mathcal{X}$ is Pareto optimal if it is not dominated by any other solution. Moreover, the Pareto set $\mathcal{P}$ comprises all Pareto optimal solutions, formally, $\mathcal{P} = \{\pi^* \in \mathcal{X}\mid \nexists \; \pi\in \mathcal{X}: \pi \prec \pi^* \}$. The Pareto front $\mathcal{F}$ is the visualization of Pareto optimal solutions in the objective space, i.e., $\mathcal{F} = \left\{F(\pi^*) \mid \pi^* \in \mathcal{P}\right\}$.

\noindent\textbf{Decomposition Method:} To solve MOCOPs, decomposition-based methods are widely used to decompose the original problem into a set of subproblems (i.e., SOCOPs). For all decomposition-based baselines computed in this paper, we choose Weighted-Sum (WS) as the decomposition technique following prior studies \cite{WE,POCCO}. Specifically, given a MOCOP, the objective of the $s$th subproblem is defined by the weight vector $\lambda_s$ as  $\min_{\pi \in \mathcal{X}} \sum\nolimits_{j=1}^{\kappa}\lambda_{s}^{j} F^{j}(\pi)$.
 Notably, effectively leveraging $\lambda_s$ is crucial for improving the performance of the neural solver.
 \section{The Decoder Architecture of WeCon-CCO}
\label{cco}

\begin{figure}[!t]
    \centering
    \includegraphics[width=1\linewidth]{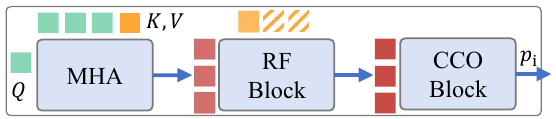}
    \caption{Illustration of WeCon-CCO decoder architecture.}
    \label{fig:WeCono}
\end{figure}

Existing neural solvers typically rely on a single network with limited capacity to handle all subproblems, which can overcomplicate learning and lead to suboptimal performance. Accordingly, several studies have explored MoE-based solvers that route subproblems to specialized experts \cite{mvmoe,SHIELD}. A representative work is \cite{POCCO}, which proposes a decoder consisting of an MHA layer followed by an MoE-based CCO block tailored for solving MOCOPs.   To demonstrate the generality of the proposed RF block, we adopt the decoder from \cite{POCCO} and insert RF between its MHA layer and CCO block, yielding the variant WeCon-CCO. 
Formally, for WeCon-CCO, the vector ${\bm{q}_c^t}$ is computed  as follows:
\begin{equation}
\hat{\bm{H}}_1^t =  \operatorname{MHA}\left(\bm{h}_q^t, [\bm{H}||\bm{A}], [\bm{H}||\bm{A}] \right),
\end{equation}
\begin{equation}
\footnotesize   
\hat{\bm{H}}_2^t=\hat{\bm{H}}_1^t +\Big( \big(\mathrm{ReLU}([\hat{\bm{H}}_1^t||\bm{A}_{\rightarrow n}]\bm{W}_6+\bm{b}_6)\big)\bm{W}_7 +\bm{b}_7 \Big  
),
\end{equation}
\begin{equation}
 \bm{q}_c^t =\operatorname{RN} \left(\sum\nolimits_{i=1}^{E} G_i(\hat{\bm{H}}_2^t )E_i(\hat{\bm{H}}_2^t)+\hat{\bm{H}}_2^t\right),
\label{eq:cco}
\end{equation}
where $E$ denotes the number of experts, $G_i$ and $E_i$ denote the output of the $i$th gate and expert functions, respectively. The WeCon-CCO's decoder architecture is presented in Figure~\ref{fig:WeCono}. In addition, we replace the Instance Normalization used in the original CCO with RMSNorm to remain consistent with the normalization employed in the WeCon encoder.
\section{EPO Procedures}
\label{alg:epo}
As shown in Algorithm~\ref{alg:algorithm}, we present the pseudocode of EPO, which  efficiently generates sufficiently high-quality solutions
and thereby constructs more informative preference pairs.
\begin{algorithm}[!t]
\caption{EPO Procedures}
\label{alg:algorithm}
\begin{algorithmic}[1]
\STATE \textbf{Input:} Instance distribution $\tilde{{G}}$, weight vector distribution $\tilde{\lambda}$, training steps $E$, batch size $B$, the total number of sampled solutions $r$, and guidance ratio parameter $c$. 
\STATE \textbf{Output:} Trained policy network parameters $\theta$.
\STATE Initialize policy network parameters $\theta$.
\FOR{$e = 1$ to $E$}
    \FOR{$b = 1$ to $B$}
   \STATE{$\lambda_{b} \sim $ \textsc{SampleWeightVector}($\tilde{\lambda}$)}
   
   \STATE{${G}_b \sim $ \textsc{SampleInstance} ($\tilde{G}$)}

\STATE $\pi_{i,b}\sim \textsc{GuidedSampleSolution}(p_\theta(\cdot\mid {G}_b,\lambda_b)), $ \quad $ \forall i \in \{\,1+mc \mid m\in\mathbb{Z}_{\ge 0},\ 1+mc \le r\,\}$  

    \STATE{$\pi_{i,b}\! \sim $\! \textsc{RandomSampleSolution}($p_{\theta}(\cdot\mid {G}_b, \lambda_b)$), $\forall i\in\{1,...,r\}\setminus \{\,1+mc \mid m\in\mathbb{Z}_{\ge 0},\ 1+mc \le r\,\}$}
        
\STATE{$y(\pi_{i,b},\pi_{j,b}) \leftarrow$ \textsc{PairwisePreference}$(1_{[\pi_{i,b}\lessdot \pi_{j,b}]})$, \quad $\forall\, (i,j) \in \{(i,j)\mid 1 \leq i < j \leq r\} $}, 
\ENDFOR
\STATE{Compute gradient $\nabla_{\theta} \mathcal{L}(\theta)$ according to Eq.~(\ref{eq:loss})}
        
\STATE{$\theta \leftarrow$ \textsc{Adam}($\theta, \nabla_{\theta} \mathcal{L}(\theta)$)}
\ENDFOR
\end{algorithmic}
\end{algorithm}

\section{Experimental Setups}
\label{app:setup}
This section describes the hyperparameter settings and hardware setup, as well as the definitions, context embeddings construction, instance augmentation method, and test datasets for the four MOCOP variants, respectively.
 
\noindent {\textbf{Hyper-parameters:}  For the method-specific hyperparameters introduced by WeCon, we set $k=5$, $c=8$, $r=64$, and $C=10$ in all experiments conducted in this paper. In addition, we configure most hyperparameters based on the prior work \cite{WE}.  Following the prior studies \cite{POCCO}, for a fair comparison,  we set $\beta=3.5$ for bi-objective problems and $\beta=4.5$ for tri-objective problems. In addition, all solvers are trained for 200 epochs, with each epoch processing 100,000 randomly sampled instances and a batch size of $B = 64$. We use the Adam optimizer~\cite{kingma2014adam} with a learning rate of $3 \times 10^{-4}$ and a weight decay of $10^{-6}$. The $\mathcal{N}$ weight vectors used for decomposition are generated following~\cite{das1996normal}, with $\mathcal{N} = 101$ for $\kappa = 2$ and $\mathcal{N} = 105$ for $\kappa = 3$. To ensure a fair comparison, we follow \cite{POCCO} when setting the reference and ideal points; the corresponding values are reported in Table~\ref{point}.

\begin{table}[!t]
\centering
\caption{Reference points and ideal points}
\vspace{-0.2cm}
\label{point}
\begin{tabular}{@{} l c c c @{}}
\toprule
Problem   & Size & $r^*$               & $z$       \\ 
\midrule
\multirow{7}{*}{Bi-TSP}
          & 20  & (20,\,20)         & (0,\,0)   \\
          & 50  & (35,\,35)         & (0,\,0)   \\
          & 100 & (65,\,65)         & (0,\,0)   \\
          & 150 & (85,\,85)         & (0,\,0)   \\
          & 200 & (115,\,115)       & (0,\,0)   \\
          & 500 & (250,\,250)       & (0,\,0)   \\
          & 1000 & (450,\,450)       & (0,\,0)   \\
\midrule
\multirow{3}{*}{Bi-CVRP}
          & 20  & (30,\,4)          & (0,\,0)   \\
          & 50  & (45,\,4)          & (0,\,0)   \\
          & 100 & (80,\,4)          & (0,\,0)   \\
\midrule
\multirow{5}{*}{Bi-KP}
          & 50  & (5,\,5)           & (30,\,30) \\
          & 100 & (20,\,20)         & (50,\,50) \\
          & 200 & (30,\,30)         & (75,\,75) \\
          & 500 & (90,\,90)         & (150,\,150) \\
          & 1000 & (130,\,130)         & (260,\,260) \\
\midrule
\multirow{3}{*}{Tri-TSP}
          & 20  & (20,\,20,\,20)    & (0,\,0)   \\
          & 50  & (35,\,35,\,35)    & (0,\,0)   \\
          & 100 & (65,\,65,\,65)    & (0,\,0)   \\
\bottomrule
\end{tabular}
\end{table}

\noindent \textbf{Hardware:} We comprehensively evaluate the performance of all algorithms, using a machine equipped with an Intel Xeon Gold 6348 and an NVIDIA A800 GPU (80GB).

\noindent\textbf{MOTSP:}  An MOTSP (Bi-TSP/Tri-TSP) instance is defined by multiple cost matrices and aims to identify a set of tours (i.e., node sequences) that are Pareto optimal. Specifically, a $\kappa$-objective TSP instance ${G}$ with $n$ nodes is characterized by cost matrices $C^{i} = (c_{j,k}^{i})$, where $i \in \left\{ 1, \cdots, \kappa \right\}$ and $j,k \in \left\{ 1, \cdots, n \right\}$. The $\kappa$ objectives are defined as follows:
\begin{equation}
\begin{split}
    &\min_{\pi\in \mathcal{X}} F(\pi) = \min (F^{1}(\pi), F^{2}(\pi), \cdots, F^{\kappa}(\pi)), \\
    &{\rm with}\; F^{i}(\pi) = c_{\pi_{n},\pi_{0}}^{i} + \sum\nolimits_{j=1}^{n-1}c_{\pi_{j},\pi_{j+1}}^{i},
\end{split}   
\end{equation}
where $\pi=\left(\pi_{1}, \pi_{2}, \cdots, \pi_{n}\right)$ with $\pi_{j}\in\left\{1, \cdots, n\right\}$. $\mathcal{X}$ denotes the set of all feasible solutions (i.e., tours), where each node is visited exactly once. Following \cite{Li_deep_2021,lin_pareto_2022}, we consider the Euclidean MOTSP. Each node $j$ is associated with a $2\kappa$-dimensional feature vector $o_j=[loc_{j}^{1}, loc_{j}^{2}, \cdots, loc_{j}^{\kappa}]$, where $loc_{j}^{i} \in \mathbb{R}^{2}$ denotes the coordinate under the $i$-th objective. The $i$-th objective value is computed as $F_{i}(\pi)=||loc_{\pi_{n}}^{i}-loc_{\pi_{1}}^{i}||_{2}+\sum_{j=1}^{n-1}||loc_{\pi_{j}}^{i}-loc_{\pi_{j+1}}^{i}||_{2}$.

\noindent\textbf{Bi-CVRP:} A Bi-CVRP instance consists of $n$ customer nodes and a single depot node. Each node $j\in\left\{0,\cdots,n\right\}$ is associated with a 3-dimensional feature vector $o_j=[loc_j,\delta_j]$, where $loc_j$ and $\delta_j$ denote the coordinates and demand of node $j$, respectively. For the depot, the demand is set to zero, i.e., $\delta_0=0$. A fleet of vehicles with capacity $\mathcal{Q}$ (with $\mathcal{Q}>\delta_j$ for all customers) is used to serve all customers via multiple routes, each starting and ending at the depot. The solution must satisfy the following constraints: (i) each customer is visited exactly once, and (ii) the total demand served on each route does not exceed the vehicle capacity.
In this paper, following prior studies \cite{lin_pareto_2022}, we aim to minimize two objectives: i.e., the total length of all routes and the length of the longest route.

\noindent\textbf{Bi-KP.} 
A Bi-KP instance consists of $n+1$ items, where each item $j \in \{1,\cdots,n\}$ is characterized by a 2-dimensional feature vector $o_j=[w_j, u_j]$, with $w_j$ denoting its weight and $u_j$ its profit vector. Here, $u_j \in \mathbb{R}^2$ contains the two objective-specific profit values associated with item $j$. We consider a knapsack with capacity $\mathcal{C}$, where $\mathcal{C}>w_j$ for each item. The goal is to select a subset of items such that the total weight does not exceed $\mathcal{C}$. Following the prior studies \cite{lust2010two,zhou2012multiobjective}, we aim to simultaneously maximize two objectives: the sum of the two profit components across the selected items.

\begin{table*}[!t]
  \centering
  \caption{Performance comparison of different methods on Bi-CVRP across different problem scales}
  \label{tab:bicvrp}
\vspace{-0.2cm}
      \begin{tabular}{lccc ccc ccc}
        \toprule
         \multirow{2}{*}{Method} & \multicolumn{3}{c}{Bi-CVRP20}
          & \multicolumn{3}{c}{Bi-CVRP50}
          & \multicolumn{3}{c}{Bi-CVRP100} \\
          \cmidrule(lr){2-4}\cmidrule(lr){5-7}\cmidrule(lr){8-10}
           & HV$\uparrow$     & Gap$\downarrow$     & Time$\downarrow$    & HV$\uparrow$     & Gap$\downarrow$      & Time$\downarrow$    & HV$\uparrow$     & Gap$\downarrow$       & Time$\downarrow$   \\
        \midrule
      NSGA-II  (TEVC'02)    & 0.4275 & 0.63\%   & 6.4h   & 0.3896 &5.21\%   & 8.8h   & 0.3620 & 11.49\%   & 9.4h   \\
      MOGLS  (EJOR'02)     & 0.4278 & 0.56\%   & 9.0h   & 0.3984 & 3.07\%   & 20h    & 0.3875 & 5.26\%    & 72h    \\
    MOEA/D  (EJOR'07)    & 0.4255 & 1.09\%   & 2.3h   & 0.4000 & 2.68\%   & 2.9h   & 0.3953 & 3.35\%    & 5.0h   \\
      PPLS/D-C  (TCYB'24)     & 0.4287 & 0.35\%   & 1.6h   & 0.4007 & 2.51\%   & 9.7h   & 0.3946 & 3.52\%    & 38h    \\
      \midrule
      DRL-MOA  (TCYB'21)     & 0.4287 & 0.35\%   & 8s     & 0.4076 & 0.83\%   & 12s    & 0.4055 & 0.86\%    & 21s    \\
      MDRL  (TNNLS'23)       & 0.4291 & 0.26\%   & 6s     & 0.4082 & 0.68\%   & 13s    & 0.4056 & 0.83\%    & 22s    \\ 
   \rowcolor{lightgray} MDRL-A  (TNNLS'23)   & 0.4294 & 0.19\%   & 12s    & 0.4092 & 0.44\%   & 36s    & 0.4072 & 0.44\%    & 2.8m   \\
      EMNH  (NeurIPS'23)       & 0.4299 & 0.07\%   & 7s     & 0.4098 & 0.29\%   & 12s    & 0.4072 & 0.44\%    & 22s    \\ 
  \rowcolor{lightgray}  EMNH-A  (NeurIPS'23)  & \underline{0.4302} & \underline{0.00\%}  & 12s    & 0.4106 & 0.10\%   & 35s    & 0.4079 & 0.27\%    & 2.8m   \\
\midrule
      PMOCO  (ICLR'22)     & 0.4267 & 0.81\%   & 3s     & 0.4036 & 1.80\%   & 7s    & 0.3913 & 4.33\%    & 16s    \\
  \rowcolor{lightgray}    PMOCO-A (ICLR'22)  & 0.4294 & 0.19\%   & 6s    & 0.4080 & 0.73\%   & 21s    & 0.3969 & 2.96\%    & 1.6m    \\
      CNH    (TNNLS'25)     & 0.4287 & 0.35\%   & 4s    & 0.4087 & 0.56\%   & 8s    & 0.4065 & 0.61\%    & 15s    \\
  \rowcolor{lightgray}  CNH-A  (TNNLS'25)    & 0.4299 & 0.07\%   & 7s    & 0.4101 & 0.22\%   & 23s    & 0.4077 & 0.32\%    & 1.7m   \\
    WE-CA   (ICLR'25)     & 0.4290 & 0.28\%   & 3s     & 0.4088 & 0.54\%   & 7s    & 0.4069 & 0.51\%    & 15s    \\ 
  \rowcolor{lightgray}  WE-CA-A (ICLR'25)   & 0.4300 & 0.05\%   & 6s    & 0.4103 & 0.17\%   & 19s    & 0.4081 & 0.22\%    & 1.5m   \\
    PA-MoE-W (OpenReview'25)    & 0.4291& 0.26\%  &  7s   &  0.4095  & 0.36\%   & 14s  & 0.4073 & 0.42\%    &  31s  \\
   \rowcolor{lightgray} PA-MoE-W-A (OpenReview'25) & 0.4301 & 0.02\%   &  14s&  0.4106& 0.10\%   & 41s  &  0.4084 & 0.15\%    &    2.9m\\
       GF-MOCA (OpenReview'25)    & 0.4295& 0.16\%  &  8s   &  0.4092  & 0.44\%   & 17s  & 0.4073 & 0.42\%    &  46s  \\
   \rowcolor{lightgray} GF-MOCA-A (OpenReview'25) & 0.4301 & 0.02\%   &  24s&  0.4104& 0.15\%   & 1.2m &  0.4082 & 0.20\%    &    4.0m\\
    POCCO-W (NeurIPS'25)    & 0.4294 & 0.19\%   & 7s     & 0.4102 & 0.19\%   & 15s    & 0.4084 & 0.15\%    & 34s    \\
 \rowcolor{lightgray}     POCCO-W-A (NeurIPS'25) & 0.4301 & 0.02\%   & 18s    & 0.4108 & 0.05\%   & 51s   & \underline{0.4089} & \underline{0.02\%}    & 3.3m   \\ \midrule
      \textbf{WeCon} (ours)    & 0.4298 & 0.09\%   & 4s     &  0.4103& 0.17\%   &9s    & 0.4083 & 0.17\%    & 19s    \\
    \rowcolor{lightgray}  \textbf{WeCon-A} (ours)  & \textbf{0.4303} & \textbf{-0.02\%}   &8s    & \underline{0.4109} & \underline{0.02\%}   & 25s   & 0.4088 & 0.05\%    & 1.9m   \\
      \textbf{WeCon-CCO} (ours)    & 0.4299 & 0.07\%   & 8s     & 0.4105 & 0.12\%   & 15s    & 0.4085 & 0.12\%    & 35s    \\
   \rowcolor{lightgray}   \textbf{WeCon-CCO-A} (ours)  & \underline{0.4302}  & \underline{0.00\%}   & 15s    & \textbf{0.4110} & \textbf{0.00\%}   & 47s   &\textbf{0.4090}  & \textbf{0.00\%}    & 3.2m   \\
        \bottomrule
      \end{tabular}
\end{table*}

\noindent \textbf{Context Embedding:} Following the prior studies \cite{WE,POCCO}, for Bi-TSP and Tri-TSP, the context embedding $\bm{h}_q$ is formed by concatenating the embeddings of the first and last visited nodes. For Bi-CVRP, $\bm{h}_q$ consists of the embedding of the last visited node together with the remaining vehicle capacity. In Bi-KP, $\bm{h}_q$ combines the graph embedding $\widetilde{h} = \frac{1}{n+1}\sum_{i=0}^n h_i$
with the remaining knapsack capacity.

\noindent \textbf{Instance Augmentation:}  To further improve the inference performance of the solvers, following the prior studies \cite{lin_pareto_2022}, we apply the instance augmentation method proposed in~\cite{kwon_pomo_2020}. The rationale behind instance augmentation is that a Euclidean VRP instance can be transformed into multiple equivalent instances that share the same optimal solution, e.g., by flipping the coordinates of all nodes. Given a coordinate $(x, y)$ in a VRP, there are eight standard transformations, i.e., $(x', y')$ = $(x, y); (y, x); (x, 1-y); (y, 1-x); (1-x, y); (1-y, x)$; $(1-x, 1-y); (1-y, 1-x)$. In our paper, we adopt these transformations independently to each coordinate set.  Consequently, this yields eight transformations for Bi-CVRP (one coordinate set per node), $8^2=64$ transformations for Bi-TSP, and $8^3=512$ transformations for Tri-TSP.

\begin{table*}[!t]
\centering
\caption{Performance comparison of different methods on Bi-KP across different problem scales}
\label{tab:bikp}
\vspace{-0.2cm}
\renewcommand\arraystretch{0.5}

\begin{tabular}{l ccc ccc ccc}
\toprule
\multirow{2}{*}{Method}
  & \multicolumn{3}{c}{Bi-KP50} & \multicolumn{3}{c}{Bi-KP100} & \multicolumn{3}{c}{Bi-KP200} \\
\cmidrule(lr){2-4} \cmidrule(lr){5-7} \cmidrule(lr){8-10}
  & HV$\uparrow$ & Gap$\downarrow$ & Time$\downarrow$ & HV$\uparrow$ & Gap$\downarrow$ & Time$\downarrow$ & HV$\uparrow$ & Gap$\downarrow$ & Time$\downarrow$ \\
\midrule
      WS-DP      & \textbf{0.3561}  & \textbf{0.00\%}   & 22m    & 0.4532  & 0.02\%   & 2h     & 0.3601  & 0.03\%   & 5.8h   \\
      NSGA-II  (TEVC'02)    & 0.3547  & 0.39\%   & 7.8h   & 0.4520  & 0.29\%   & 8.0h   & 0.3590  & 0.33\%   & 8.4h   \\
      MOGLS  (EJOR'02)    & 0.3540  & 0.59\%   & 5.8h   & 0.4510  & 0.51\%   & 10h    & 0.3582  & 0.56\%   & 18h    \\
    MOEA/D (TEVC'07)    & 0.3540  & 0.59\%   & 1.6h   & 0.4508  & 0.55\%   & 1.7h   & 0.3581  & 0.58\%   & 1.8h   \\
      PPLS/D-C  (TCYB'24)   & 0.3528  & 0.93\%   & 18m    & 0.4480  & 1.17\%   & 47m    & 0.3541  & 1.69\%   & 1.5h   \\
      \midrule
      DRL-MOA  (TCYB'21)   & \underline{0.3559}  & \underline{0.06\%}   & 8s     & 0.4531  & 0.04\%   & 15s    & 0.3601  & 0.03\%   & 32s    \\
      MDRL (TNNLS'23)       & 0.3530  & 0.87\%   & 7s     & 0.4532  & 0.02\%   & 18s    & 0.3601  & 0.03\%   & 35s    \\
      EMNH  (NeurIPS'23)      & \textbf{0.3561}  & \textbf{0.00\%}   & 7s     & \textbf{0.4535}  & \textbf{-0.04\%}  & 17s    & \textbf{0.3603}  & \textbf{-0.03\%}   & 48s    
      \\ \midrule
      PMOCO  (ICLR'22)     & 0.3552  & 0.25\%   & 5s     & 0.4523  & 0.22\%   & 9s    & 0.3595  & 0.19\%   & 35s    \\
      CNH  (TNNLS'25)       & 0.3556  & 0.14\%   & 5s    & 0.4527  & 0.13\%   & 10s    & 0.3598  & 0.11\%   & 35s    \\
      WE-CA (ICLR'25)      & 0.3558  & 0.08\%   & 5s     & 0.4531  & 0.04\%   & 10s    & \underline{0.3602}  & 0.00\%   & 34s    \\
    PA-MoE-W (OpenReview'25)    & 0.3507 & 1.52\%  &  10s   &  0.4430 & 2.27\%   & 21s  & 0.3309 & 8.13\%    &  53s  \\
        GF-MOCA (OpenReview'25)    & 0.3560 & 0.00\%  &  11s   &  0.4534 & -0.02\%   & 26s  & 0.3602& 0.00\%    &  2m  \\
      POCCO-W (NeurIPS'25)   & \textbf{0.3561}  & \textbf{0.00\%}   & 11s    & \underline{0.4534}  & \underline{-0.02\%}   & 23s    & \textbf{0.3603}  & \textbf{-0.03\%}   & 1.0m   \\ \midrule
      \textbf{WeCon} (ours)    & 0.3558  & 0.08\%   &5s    & 0.4531  & 0.04\%   & 10s    & \underline{0.3602}  & \underline{0.00\%}   & 35s \\
      \textbf{WeCon-CCO} (ours)    & \textbf{0.3561}  & \textbf{0.00\%}   & 11s    & 0.4533 & 0.00\%   & 23s    & \underline{0.3602}  & \underline{0.00\%}   & 1.0m 
\\ \bottomrule
      \end{tabular}
\end{table*}    
\noindent \textbf{Test Datasets:} Except for Bi-TSP500, Bi-TSP1000, Bi-KP500, and Bi-KP1000, all other test datasets are taken from the prior study \cite{WE}. Because \cite{WE} did not conduct experiments on large-scale instances, we generate the test datasets for Bi-TSP500, Bi-TSP1000, Bi-KP500, and Bi-KP1000 by sampling instances under a uniform distribution, with 20 instances for each size.

 \begin{table*}[!t]
\centering
\caption{Performance comparison of different methods on Tri-TSP across different problem scales}
\label{tab:tritsp}
\vspace{-0.2cm}
\renewcommand\arraystretch{0.5}
      \begin{tabular}{lccc ccc ccc}
        \toprule
    \multirow{2}{*}{Method} & \multicolumn{3}{c}{Tri-TSP20}
      & \multicolumn{3}{c}{Tri-TSP50}
      & \multicolumn{3}{c}{Tri-TSP100} \\
      \cmidrule(lr){2-4}\cmidrule(lr){5-7}\cmidrule(lr){8-10}
         & HV$\uparrow$      & Gap$\downarrow$       & Time$\downarrow$   & HV$\uparrow$      & Gap$\downarrow$       & Time$\downarrow$   & HV$\uparrow$      & Gap$\downarrow$        & Time$\downarrow$   \\
\midrule
    WS-LKH     & \textbf{0.4712} & \textbf{0.00\%}   & 12m   & \textbf{0.4440} & \textbf{-0.11\%}  & 1.9h  & \textbf{0.5076} & \textbf{-0.59\%}   & 6.6h   \\
      NSGA-II  (TEVC'02)  & 0.4238 & 10.06\%  & 7.1h  & 0.2858 & 35.56\%  & 7.5h  & 0.2824 & 44.03\%   & 9.0h   \\
      MOGLS  (EJOR'02)    & 0.4701 & 0.23\%   & 1.5h  & 0.4211 & 5.05\%   & 4.1h  & 0.4254 & 15.70\%   & 13h    \\
    MOEA/D  (TEVC'07)   & 0.4702 & 0.21\%   & 1.9h  & 0.4314 & 2.73\%   & 2.2h  & 0.4511 & 10.60\%   & 2.4h   \\
      PPLS/D-C (TCYB'24)  & 0.4698 & 0.30\%   & 1.4h  & 0.4174 & 5.89\%   & 3.9h  & 0.4376 & 13.28\%   & 14h    \\
      \midrule
      DRL-MOA  (TCYB'21)   & 0.4699 & 0.28\%   & 6s    & 0.4303 & 2.98\%   & 9s    & 0.4806 & 4.76\%    & 18s    \\
      MDRL (TNNLS'23)       & 0.4699 & 0.28\%   & 5s    & 0.4317 & 2.66\%   & 10s   & 0.4852 & 3.84\%    & 17s    \\
   \rowcolor{lightgray} MDRL-A  (TNNLS'23)  & \textbf{0.4712} & \textbf{0.00\%}   & 4.2m  & 0.4408 & 0.61\%   & 25m   & 0.4958 & 1.74\%    & 1.6h   \\
      EMNH  (NeurIPS'23)      & 0.4699 & 0.28\%   & 5s    & 0.4324 & 2.50\%   & 10s   & 0.4866 & 3.57\%    & 17s    \\
 \rowcolor{lightgray}   EMNH-A  (NeurIPS'23)  & \textbf{0.4712} & \textbf{0.00\%}   & 4.2m  & 0.4418 & 0.38\%   & 25m   & 0.4973 & 1.45\%    & 1.6h   \\
\midrule
    PMOCO   (ICLR'22)    & 0.4693 & 0.40\%   & 3s    & 0.4315 &  2.71\%   & 6s   & 0.4858 & 3.73\%    & 12s   \\
  \rowcolor{lightgray}  PMOCO-A  (ICLR'22)  & \textbf{0.4712} & \textbf{0.00\%}   & 4.9m  & 0.4409 & 0.59\%   & 22m   & 0.4956 & 1.78\%    & 1.6h   \\
    CNH    (TNNLS'25)      & 0.4698 & 0.30\%   & 4s  & 0.4358 & 1.74\%   & 6s   & 0.4931 & 2.28\%    & 14s    \\
 \rowcolor{lightgray}   CNH-A  (TNNLS'25)    & 0.4704 & 0.17\%   & 5.3m  & 0.4409 & 0.59\%   & 24m   & 0.4996 & 0.99\%    & 1.6h   \\
    WE-CA   (ICLR'25)     & 0.4707 & 0.11\%   & 2s    & 0.4389 & 1.04\%   & 5s    & 0.4975 & 1.41\%    & 11s    \\
\rowcolor{lightgray}    WE-CA-A (ICLR'25)   & \textbf{0.4712} & \textbf{0.00\%}   & 4.8m  & 0.4432 & 0.07\%   & 20m   & 0.5032 & 0.28\%    & 1.3h   \\
    PA-MoE-W (OpenReview'25)    & 0.4709 & 0.06\%  &  6s   &  0.4391 & 0.99\%   & 12s  & 0.4975 & 1.41\%    & 26s  \\
 \rowcolor{lightgray}  PA-MoE-W-A (OpenReview'25) & \textbf{0.4712} & 0.00\%   & 9.5m & 0.4432 & 0.07\%   & 38m  &  0.5034 & 0.24\%    &  2.7h \\

     GF-MOCA (OpenReview'25)    & 0.4712 & 0.00\%  &  6s   &  0.4393 & 0.95\%   & 13s  & 0.4979 & 1.33\%    & 23s  \\
 \rowcolor{lightgray}  GF-MOCA-A (OpenReview'25) & \textbf{0.4712} & 0.00\%   & 8.9m & 0.4432 & 0.07\%   & 33m  &  0.5031 & 0.30\%    &  2.0h \\
 
    POCCO-W  (NeurIPS'25)   & 0.4710 & 0.04\%   & 5s    & 0.4403 & 0.72\%   & 13s   & 0.4985 & 1.21\%    & 28s    \\
 \rowcolor{lightgray}   POCCO-W-A (NeurIPS'25)   & \textbf{0.4712} & \textbf{0.00\%}   & 12m  & \underline{0.4437} & \underline{-0.05\%}   & 44m   & 0.5048 & -0.04\%    & 2.9h  \\ \midrule
    \textbf{WeCon} (ours)    & \underline{0.4711} & \underline{0.02\%}   & 3s  &  0.4413 & 0.50\%   & 6s   & 0.5005 & 0.81\%    & 15s \\
   \rowcolor{lightgray} \textbf{WeCon-A} (ours)    & \textbf{0.4712} & \textbf{0.00\%}   & 5.6m  & \underline{0.4437} & \underline{-0.05\%}   & 25m   & \underline{0.5049} & \underline{-0.06\%}    &  1.7h   \\
        \textbf{WeCon-CCO} (ours)    & \underline{0.4711} & \underline{0.02\%}   & 6s  & 0.4411 & 0.54\%   & 12s   & 0.5001 & 0.89\%    & 30s  \\
 \rowcolor{lightgray}   \textbf{WeCon-CCO-A} (ours)    & \textbf{0.4712} & \textbf{0.00\%}   & 8.3m  & 0.4435 & 0.00\%   & 42m   & 0.5046  & 0.00\%    & 2.9h  \\
\bottomrule
\end{tabular}
\end{table*}

\section{Results on Other MOCOP Instances}
\label{otherMOCOPS}
 In this section, following the prior studies \cite{WE,POCCO}, we assess the performance of WeCon and WeCon-CCO on other MOCOPs, including Bi-CVRP, Bi-KP, and Tri-TSP. As shown in Tables~\ref{tab:bicvrp}$\sim$\ref{tab:tritsp}, WeCon and WeCon-CCO achieve superior performance across different problem scales, demonstrating their generality beyond Bi-TSP. These results further demonstrate the effectiveness of our encoder architecture, RF module, and EPO strategy.

\section{Results on Real-world MOCOP Instances}
\label{sec5.3}

 \begin{figure*}[!t]
\centering
\subfloat{
\includegraphics[width=0.33\textwidth]{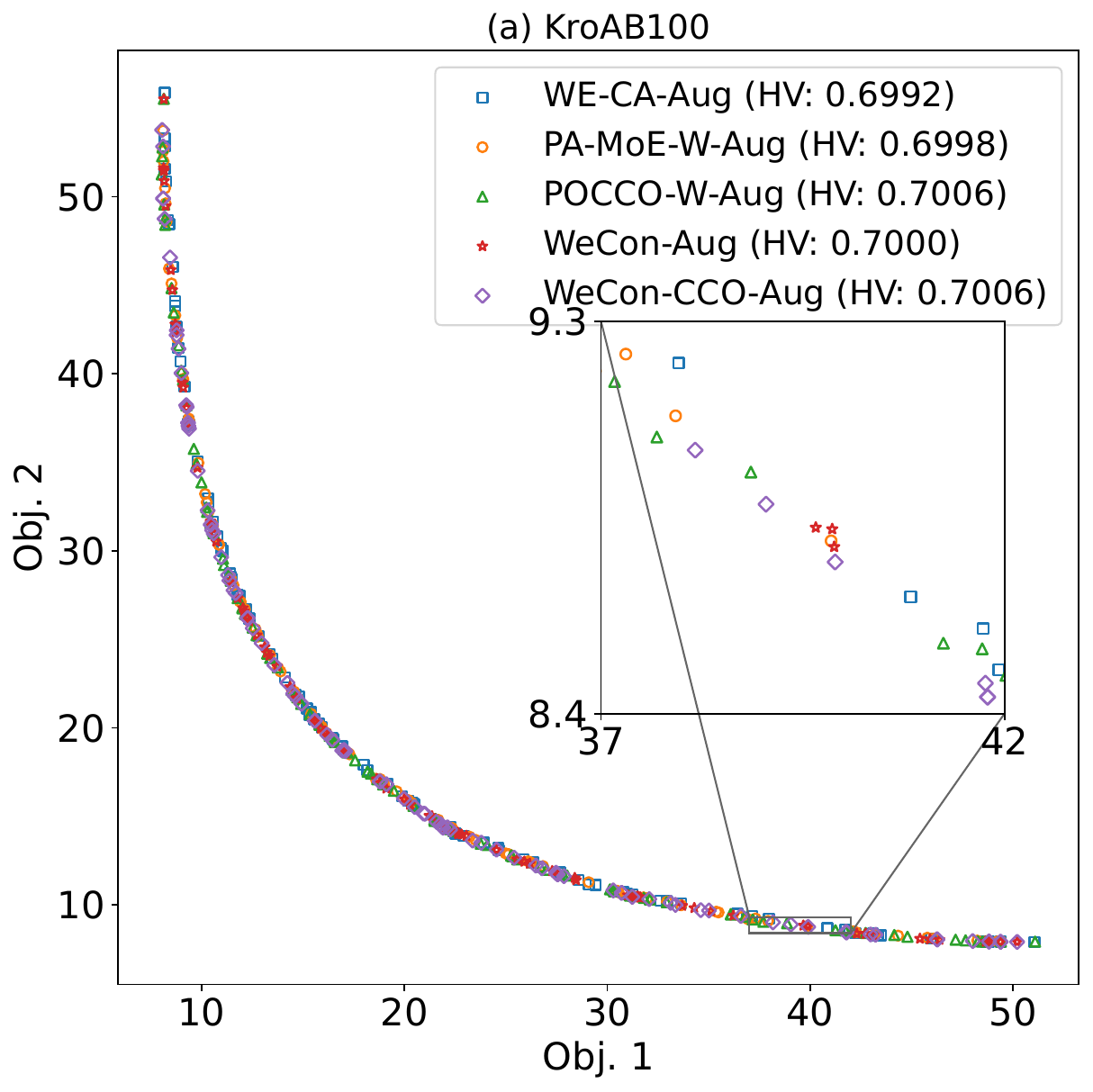}
}
\subfloat{
\includegraphics[width=0.33\textwidth]{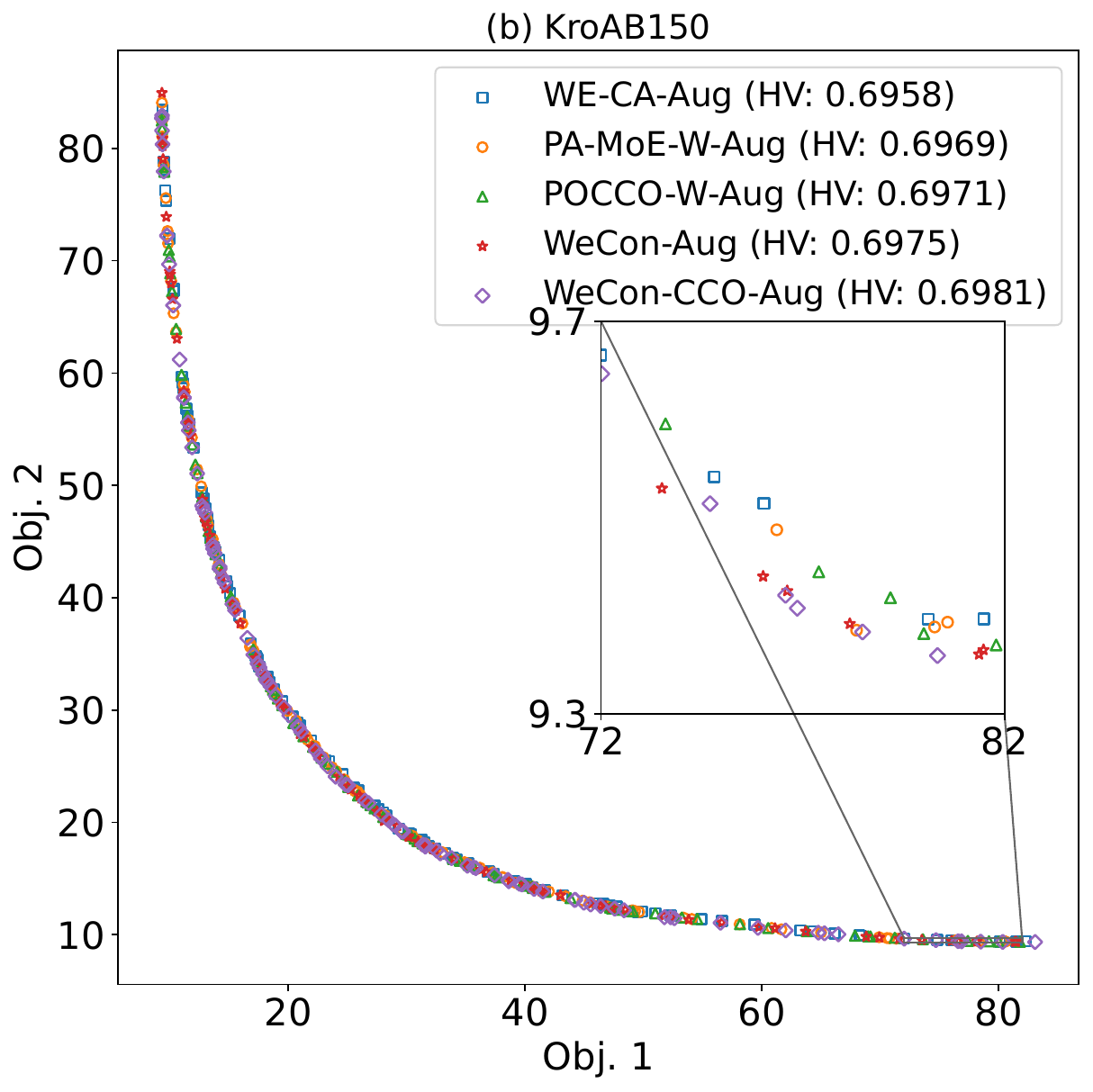}
}
\subfloat{
\includegraphics[width=0.33\textwidth]{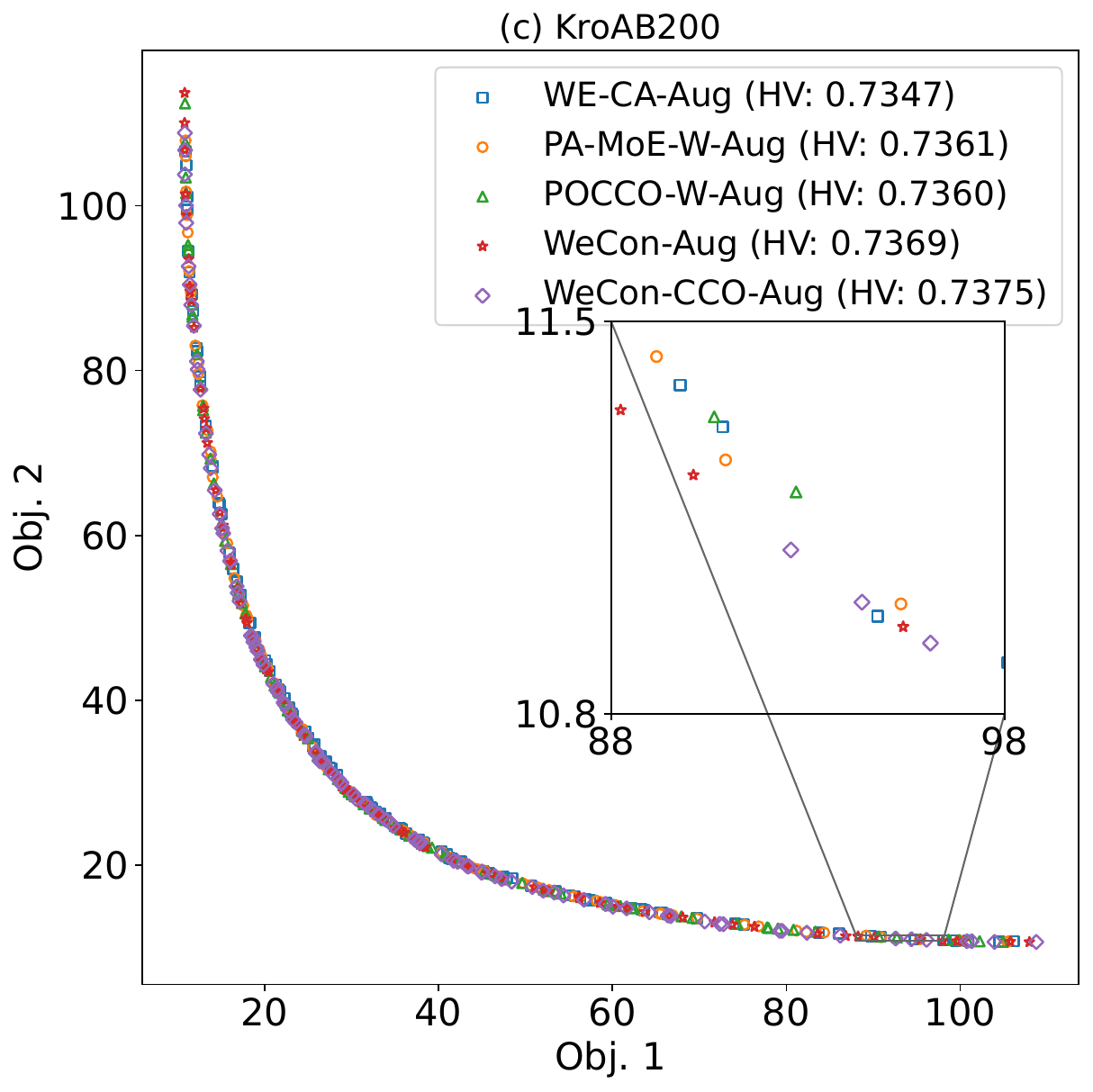}
}
\caption{Pareto fronts of benchmark instances.}
  \label{fig:vis}
  \vspace{-0.2cm}
\end{figure*}

In this section, we visualize the Pareto fronts produced by different methods on KroAB100, KroAB150, and KroAB200 instances.  As shown in Figure~\ref{fig:vis}, many solutions generated by WeCon-CCO dominate those of WE-CA, PA-MoE-W, and POCCO-W, indicating high-level performance on the associated decomposed subproblems.

\section{Results on Large-scale MOCOP Instances}
\label{sec5.2}
\begin{table*}[!t]
  \centering
  \caption{Performance comparison on large-scale instances.}
  \vspace{-0.2cm}
  \label{large_size}
  \begin{tabular}{l|cc cc cc cc}
    \toprule
    \multirow{2}{*}{Method}
    & \multicolumn{2}{c}{Bi-TSP500}
    & \multicolumn{2}{c}{Bi-TSP1000}
    & \multicolumn{2}{c}{Bi-KP500}
    & \multicolumn{2}{c}{Bi-KP1000} \\
    \cmidrule(lr){2-3}
    \cmidrule(lr){4-5}
    \cmidrule(lr){6-7}
    \cmidrule(lr){8-9}
    & HV$\uparrow$ & Time$\downarrow$
    & HV$\uparrow$ & Time$\downarrow$
    & HV$\uparrow$ & Time$\downarrow$
    & HV$\uparrow$ & Time$\downarrow$ \\
    \midrule
    WE-CA
    & 0.7335 & 1.3m
    & 0.6833 & 9.2m
    & 0.6497 & 41s
    & 0.3166 & 4.0m \\

    PA-MoE-W
    & 0.7612 & 2.6m
    & 0.7489 & 15m
    & 0.2704 & 1.2m
    & 0.2334 & 4.8m \\

    POCCO-W
    & 0.7442 & 2.2m
    & 0.7037 & 12m
    & 0.6577 & 1.1m
    & \underline{0.5733} & 4.3m \\

    WeCon
    & \underline{0.7664} & 1.4m
    & \underline{0.7733} & 9.6m
    & \textbf{0.8657} & 45s
    & \textbf{0.7778} & 4.2m \\

    WeCon-CCO
    & \textbf{0.7672} & 2.3m
    & \textbf{0.7735} & 12m
    & \underline{0.6718} & 1.2m
    & 0.5225 & 4.3m \\
    \bottomrule
  \end{tabular}
\end{table*}

In this section, following the prior art \cite{POCCO}, we assess the performance of WeCon and \mbox{WeCon-CCO} on large-scale MOCOP instances, including Bi-TSP500, Bi-TSP1000, Bi-KP500, and Bi-KP1000. As shown in Table~\ref{large_size}, WeCon-CCO achieves the best performance on the highly challenging Bi-TSP500 and Bi-TSP1000 instances, followed by WeCon. On large-scale Bi-KP instances, most methods exhibit substantial performance degradation, whereas WeCon attains the best performance. These findings further validate the effectiveness of our proposed encoder and decoder architectures and EPO strategy.

\section{Analyses of Weight-Signal Dilution during
Decoding}
\label{appendixH}
 \begin{figure}[!t]
\centering
\subfloat{
\includegraphics[width=0.48\textwidth]{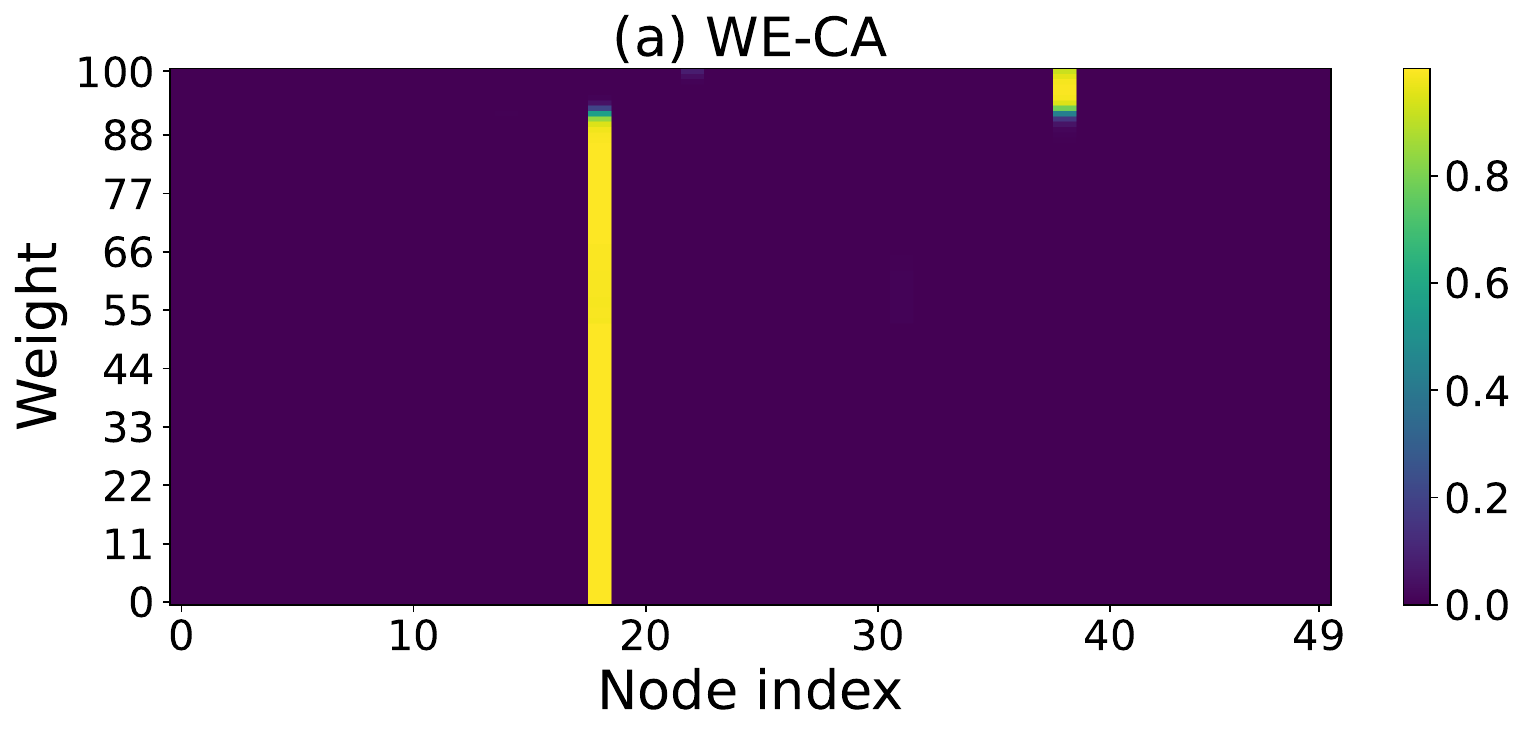}
}
\\
\subfloat{
\includegraphics[width=0.48\textwidth]{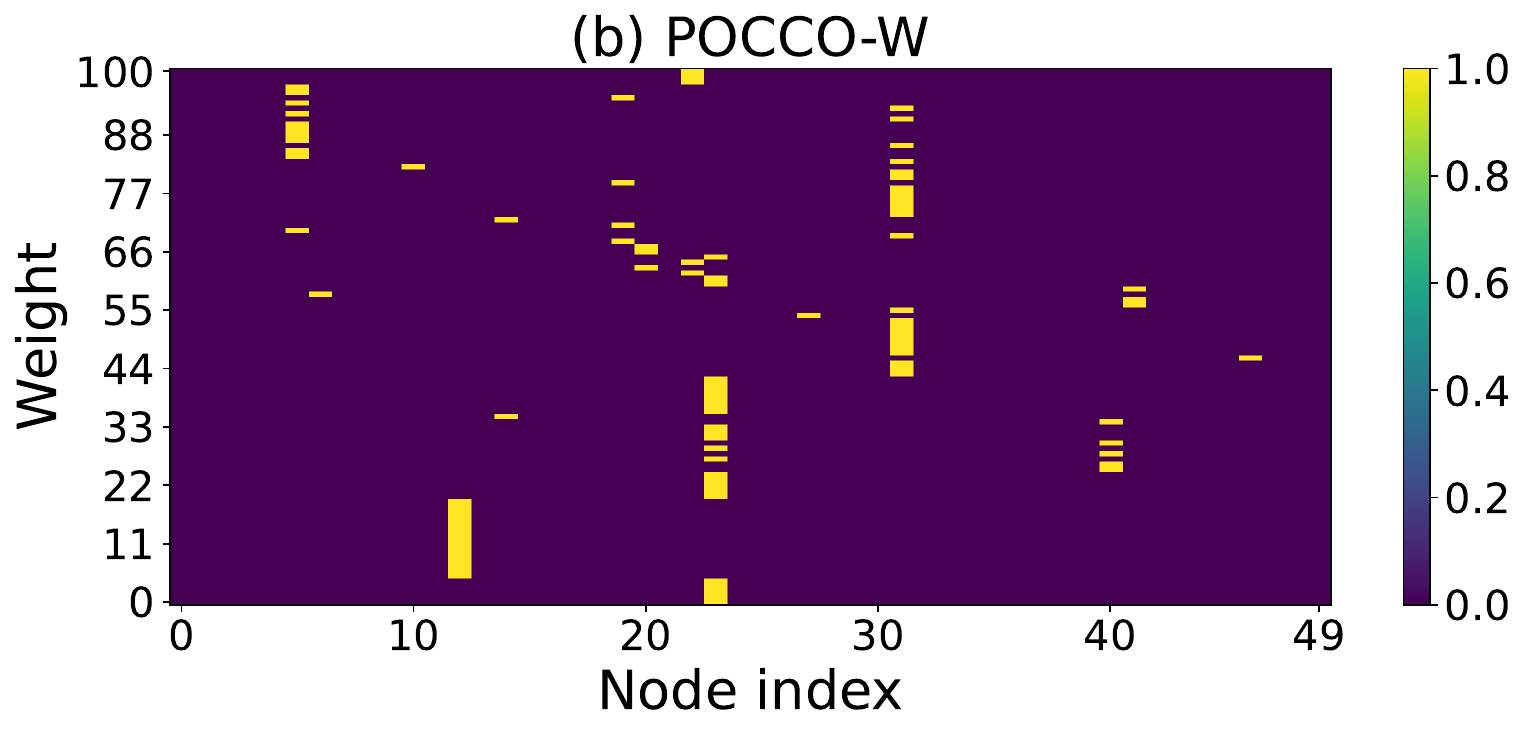}
}
\\
\subfloat{
\includegraphics[width=0.48\textwidth]{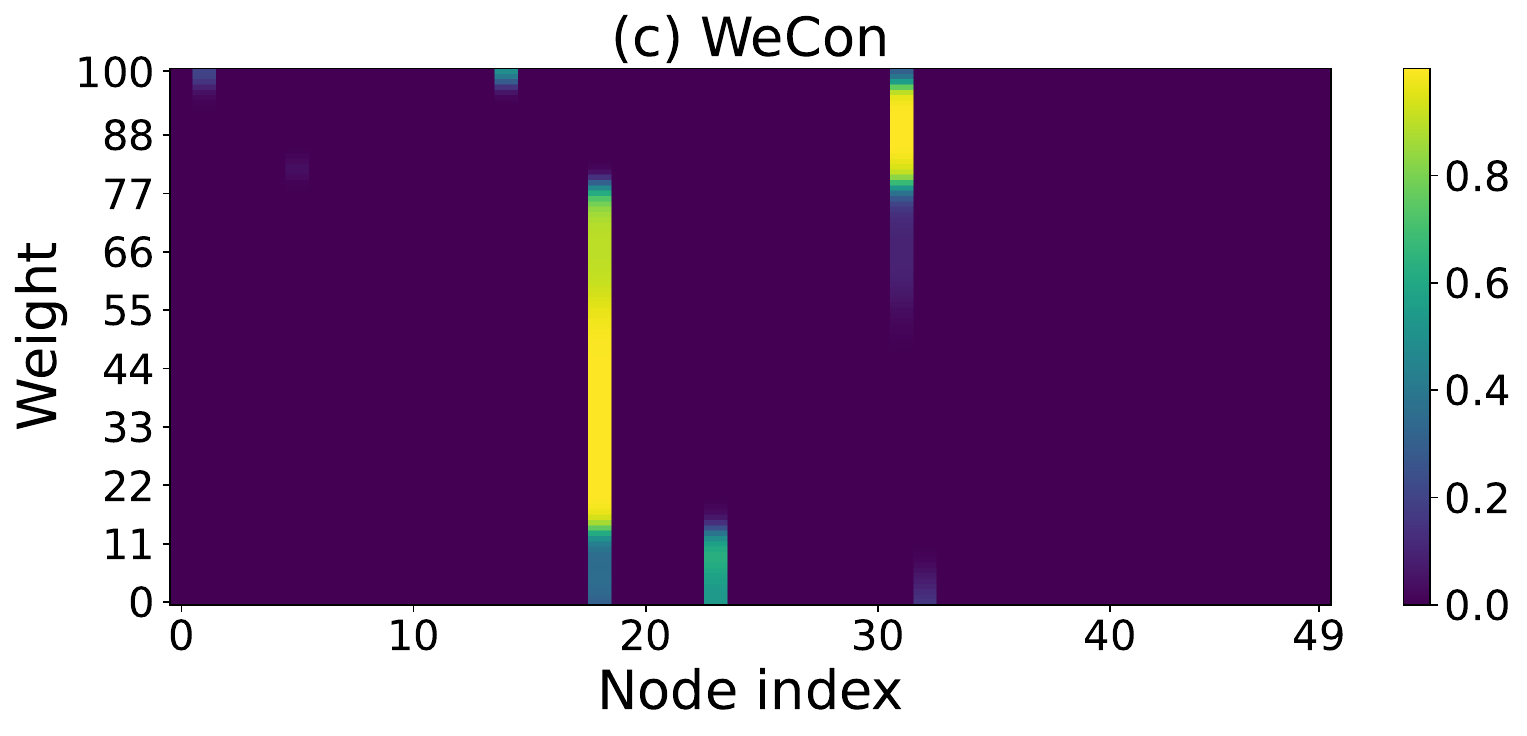}
}
\caption{selection under varying weight vectors.}
\vspace{-0.2cm}
\label{fig:decoding}
\end{figure}

In Figure~\ref{fig:decoding}, we visualize the node selection distributions of different decoders at the first decoding step on a Bi-TSP50 instance across a sweep of weight vectors, where $\mathcal{N}=101$ is used for illustration, to assess whether weight signals are diluted during decoding. As shown, WE-CA selects nearly the same nodes across weight vectors. In contrast, WeCon selects different nodes as the weight vector changes, indicating the mitigation of weight-signal dilution. Although POCCO-W exhibits greater variation due to its MoE-based
design, its node selections fluctuate irregularly across adjacent weight vectors rather than following a smooth weight-conditioned pattern. Moreover, integrating MoE incurs additional runtime, which hinders its applicability to time-sensitive MOCOPs.

\section{Model Size and Efficiency Analyses}
\label{size}
\begin{table}[!t]
\centering
\caption{Model size and training cost comparison}
\vspace{-0.2cm}
\label{para}
\begin{tabular}{l c c c}
\toprule
Method & \#Param. & \makecell{GPU\\Memory} & \makecell{Training\\Time} \\ 
\midrule
WE-CA     & 1.5M & 5926M  & 6.7h  \\
PA-MoE-W  & 4.3M & 11984M & 22.3h \\
POCCO-W   & 2.0M & 11920M & 24.5h \\
WeCon     & 5.4M & 9016M  & 9.1h  \\
WeCon-CCO & 5.9M & 14314M & 24.7h \\
\bottomrule
\end{tabular}
\end{table}

\begin{table}[!t]
    \centering
    \caption{Performance of different solvers under increased model size}
    \vspace{-0.2cm}
    \label{tab:largeparam}
        \setlength{\tabcolsep}{1mm}
    \begin{tabular}{l cc cc}
        \toprule
        \multirow{2}{*}{Method}  
        & \multicolumn{4}{c}{Bi-TSP50}
     \\
        \cmidrule(lr){2-3}\cmidrule(lr){4-5} 
        & HV$\uparrow$ & Time$\downarrow$ 
        & HV$\uparrow$ & Time$\downarrow$ \\ 
        \midrule
        WE-CA          & 0.6392 & 4s  & \cellcolor{lightgray}0.6413 & \cellcolor{lightgray}2.3m \\
        WE-CA (5.6M)   & 0.6389$\dagger$ & 6s  & \cellcolor{lightgray}0.6412$\dagger$ & \cellcolor{lightgray}2.9m \\
        WE-CA (4.9M)   & 0.6396 & 6s  & \cellcolor{lightgray}0.6414 & \cellcolor{lightgray}2.8m \\
        POCCO-W        & \textbf{0.6411} & 10s & \cellcolor{lightgray}\textbf{0.6418} & \cellcolor{lightgray}5.6m \\
        POCCO-W (6.1M) & 0.6409$\dagger$ & 14s & \cellcolor{lightgray}0.6418 & \cellcolor{lightgray}7.5m \\
        POCCO-W (4.4M) & 0.6408$\dagger$ & 14s & \cellcolor{lightgray}0.6417$\dagger$ & \cellcolor{lightgray}6.9m \\
        WeCon          & 0.6407 & 6s  & \cellcolor{lightgray}0.6415 & \cellcolor{lightgray}3m \\
        \midrule
        \multirow{2}{*}{Method}  
        & \multicolumn{4}{c}{Bi-TSP100}
      \\
        \cmidrule(lr){2-3}\cmidrule(lr){4-5} 
        & HV$\uparrow$ & Time$\downarrow$ 
        & HV$\uparrow$ & Time$\downarrow$ \\ 
        \midrule
        WE-CA          & 0.7034 & 11s & \cellcolor{lightgray}0.7066 & \cellcolor{lightgray}11m \\
        WE-CA (5.6M)   & 0.7027$\dagger$ & 13s & \cellcolor{lightgray}0.7061$\dagger$ & \cellcolor{lightgray}12m \\
        WE-CA (4.9M)   & 0.7041 & 13s & \cellcolor{lightgray}0.7069 & \cellcolor{lightgray}12m \\
        POCCO-W        & 0.7054 & 27s & \cellcolor{lightgray}\textbf{0.7077} & \cellcolor{lightgray}21m \\
        POCCO-W (6.1M) & 0.7052$\dagger$ & 30s & \cellcolor{lightgray}0.7076$\dagger$ & \cellcolor{lightgray}24m \\
        POCCO-W (4.4M) & 0.7051$\dagger$ & 30s & \cellcolor{lightgray}0.7071$\dagger$ & \cellcolor{lightgray}23m \\
        WeCon          & \textbf{0.7056} & 15s & \cellcolor{lightgray}\textbf{0.7077} & \cellcolor{lightgray}12m \\
        \bottomrule
    \end{tabular}
    {\\
    \textit{Note:} $\dagger$ denotes a performance decrease compared to the corresponding original model with fewer parameters. Results obtained with instance augmentation are highlighted in gray.}
\end{table}


In this section, we compare the number of parameters, GPU memory consumption, and training time of WE-CA, PA-MoE-W, POCCO-W, WeCon, and WeCon-CCO.  As shown in Table~\ref{para}, WeCon uses more parameters (approximately $2.5\times$ more than POCCO-W).  However, it requires only about 37.1\% and 60\% of POCCO-W’s training and inference time (see Tables~\ref{tab:bicvrp}$\sim$\ref{tab:tritsp}), respectively, and it also consumes less GPU memory. This efficiency gap is largely attributable to the MoE structure in \mbox{POCCO-W}, whose gating-and-routing process can incur additional
dispatch/aggregation operations, increasing both runtime and GPU memory consumption. In contrast, the carefully designed architecture enables WeCon to achieve high-level performance while achieving faster inference. Because the absolute parameter size of all solvers is only a few megabytes, the additional parameters introduced by WeCon are acceptable in most practical scenarios. In such settings, training and inference time are often more critical than parameter size, making WeCon particularly suitable.
Moreover, simply increasing the number of parameters does not necessarily improve HV. For example, WeCon has a similar parameter budget to PA-MoE-W, which incorporates MoE into the encoder, yet PA-MoE-W performs substantially worse than WeCon.

To further demonstrate that the effectiveness of WeCon is not merely due to having more parameters, we scale up WE-CA and POCCO-W in two ways: by increasing model depth and width. Specifically, we first expand their encoders from six to 24 layers, resulting in models with 5.6M and 6.1M parameters, respectively. We then increase their hidden dimensions, yielding models with 4.9M and 4.4M parameters, respectively. As shown in Table~\ref{tab:largeparam}, naively increasing depth even degrades the performance of both WE-CA and POCCO-W. Although increasing width improves WE-CA, its performance still falls short of WeCon, while POCCO-W shows a significant performance drop compared with its depth-scaled variant. These results further demonstrate the rationality of the encoder and decoder architecture design of WeCon.

\begin{table}[!t]
    \centering
    \caption{Sensitivity analyses of different parameter choices}\vspace{-0.2cm}
    \label{sensitivity}
    \begin{tabular}{l c c c}
        \toprule
        Setting & Bi-TSP50 & Bi-TSP100 & Bi-TSP150 \\ \toprule
        $k=4,c=8$ & 0.6406 & 0.7052 & 0.7033\\  
        $k=6,c=8$ & \textbf{0.6407} & 0.7053 & 0.7031 \\ 
        $k=5,c=6$ &  0.6404& 0.7048 & 0.7023 \\  
        $k=5,c=10$ & \textbf{0.6407}& 0.7054 & 0.7033 \\  
        $k=5,c=12$ & 0.6406& 0.7052 & 0.7031 \\   \midrule
        \makecell{ $k=5,c=8$ \\(adopted)}  & \textbf{0.6407} & \textbf{0.7056} & \textbf{0.7035}\\ \toprule
    \end{tabular}
\end{table}

\begin{table}[!t]
    \centering
    \caption{Ablation study on different decomposition techniques}\vspace{-0.2cm}
    \label{tab:ws}
    \begin{tabular}{l cc cc}
        \toprule
        \multirow{2}{*}{Method}  
        & \multicolumn{4}{c}{Bi-TSP50}
         \\
        \cmidrule(lr){2-3}\cmidrule(lr){4-5} 
        & HV$\uparrow$ & Time$\downarrow$
        & HV$\uparrow$ & Time$\downarrow$ \\ 
        \midrule
        WeCon (TCH) & 0.6391 & 6s 
        & \cellcolor{lightgray}0.6411 & \cellcolor{lightgray}3m \\
        WeCon (WS)  & \textbf{0.6407} & 6s 
        & \cellcolor{lightgray}\textbf{0.6415} & \cellcolor{lightgray}3m \\
        \midrule
        \multirow{2}{*}{Method}  
        & \multicolumn{4}{c}{Bi-TSP100}
    \\
        \cmidrule(lr){2-3}\cmidrule(lr){4-5} 
        & HV$\uparrow$ & Time$\downarrow$
        & HV$\uparrow$ & Time$\downarrow$ \\ 
        \midrule
        WeCon (TCH) & 0.7026 & 15s 
        & \cellcolor{lightgray}0.7062 & \cellcolor{lightgray}13m \\
        WeCon (WS)  & \textbf{0.7056} & 15s 
        & \cellcolor{lightgray}\textbf{0.7077} & \cellcolor{lightgray}12m \\
        \bottomrule
    \end{tabular}
    {\\
    \textit{Note:} Results obtained with instance augmentation are highlighted in gray.}
\end{table}

\section{Sensitivity analysis}
\label{kc}

 In Table~\ref{sensitivity}, we present the effect of $k$ and $c$ on the performance of WeCon. Overall, WeCon is insensitive to these hyperparameters within the tested ranges, with HV varying only marginally across different settings. The adopted configuration ($k=5,c=8$) consistently achieves the best or near-best performance across all scales.
 Hence, our predefined $k$ and $c$ are reasonable.

\section{Impact of Different Decomposition Techniques}
\label{app:ws}

In this section, we assess the impact of different decomposition techniques, namely Tchebycheff (TCH) and WS (the adopted method). 
As shown in Table~\ref{tab:ws}, WS consistently outperforms TCH, indicating that WS offers more robust performance across different problem scales. This finding is consistent with the results of prior studies \cite{WE,POCCO}.

\section{Preliminary Experiment}
\label{app:pre}
In this section, we conduct a preliminary experiment to further demonstrate the effectiveness and plug-and-play property of RF. Specifically, we integrate RF into WE-CA and evaluate whether this simple augmentation can improve its performance. As shown in Table~\ref{tab:pre}, WE-CA with RF achieves improved performance across different scales, demonstrating the effectiveness of RF in mitigating weight-signal dilution.

\begin{table}[!t]
    \centering
    \caption{The generalizability of RF}\vspace{-0.2cm}
    \label{tab:pre}
    \begin{tabular}{l cc cc}
        \toprule
        \multirow{2}{*}{Method}  
        & \multicolumn{4}{c}{Bi-TSP50}
         \\
        \cmidrule(lr){2-3}\cmidrule(lr){4-5} 
        & HV$\uparrow$ & Time$\downarrow$
        & HV$\uparrow$ & Time$\downarrow$ \\ 
        \midrule
        WE-CA & 0.6392 & 4s 
        & \cellcolor{lightgray}0.6413 & \cellcolor{lightgray}2.3m \\
       WE-CA + RF & \textbf{0.6399} & 4s 
        & \cellcolor{lightgray}\textbf{0.6413} & \cellcolor{lightgray}2.6m \\
        \midrule
        \multirow{2}{*}{Method}  
        & \multicolumn{4}{c}{Bi-TSP100}
    \\
        \cmidrule(lr){2-3}\cmidrule(lr){4-5} 
        & HV$\uparrow$ & Time$\downarrow$
        & HV$\uparrow$ & Time$\downarrow$ \\ 
        \midrule
        WE-CA & 0.7034 & 11s 
        & \cellcolor{lightgray}0.7066 & \cellcolor{lightgray}10m \\
        WE-CA + RF  & \textbf{0.7040} & 13s 
        & \cellcolor{lightgray}\textbf{0.7068} & \cellcolor{lightgray}11m \\
        \bottomrule
    \end{tabular}
    {\\
    \textit{Note:} Results obtained with instance augmentation are highlighted in gray.}
\end{table}


\end{document}